\documentclass{article}

\usepackage[main, final]{iaseai26}

\usepackage[utf8]{inputenc} 
\usepackage[T1]{fontenc}    
\usepackage{hyperref}       
\usepackage{url}            
\usepackage{booktabs}       
\usepackage{amsfonts}       
\usepackage{nicefrac}       
\usepackage{microtype}      
\usepackage{xcolor}         

\usepackage{lineno}
\usepackage{listings}
\usepackage[most]{tcolorbox}
\usepackage{siunitx}
\usepackage{xspace}
\usepackage{makecell}

\usepackage{multirow}
\usepackage{pifont} 
\usepackage{tikz}
\usepackage{collcell}

\usepackage{enumitem}
\usepackage{fancyvrb}

\newcommand*{\MinNumber}{-3.0}%
\newcommand*{\MidNumber}{0} %
\newcommand*{\MaxNumber}{3.0}%

\newtoggle{inTableHeader}
\toggletrue{inTableHeader}
\newcommand{\ApplyGradient}[1]{ \iftoggle{inTableHeader}{#1}{
        \ifdim #1 pt > \MidNumber pt
            \pgfmathsetmacro{\PercentColor}{max(min(100.0*(#1 - \MidNumber)/(\MaxNumber-\MidNumber),100.0),0.00)} %
            \hspace{-0.33em}\colorbox{blue!\PercentColor!white}{#1}
        \else
            \pgfmathsetmacro{\PercentColor}{max(min(100.0*(\MidNumber - #1)/(\MidNumber-\MinNumber),100.0),0.00)} %
            \hspace{-0.33em}\colorbox{red!\PercentColor!white}{#1}
        \fi
}}

\newcolumntype{R}{>{\collectcell\ApplyGradient}c<{\endcollectcell}}

\definecolor{darkgreen}{rgb}{0.00,0.50,0.00}
\definecolor{darkred}{rgb}{0.55, 0.0, 0.0}

\definecolor{prompt_backcolour}{HTML}{fff7de}
\definecolor{llmcolor}{HTML}{0048ff}
\definecolor{promptercolor}{HTML}{ff0800}
\newtcolorbox{blockquote}{colback=prompt_backcolour,grow to right by=-1mm,grow to left by=-1mm,boxrule=0pt,boxsep=0pt,breakable}

\newcommand{\xhdr}[1]{\vspace{1mm}\noindent{{\bf #1.}}}

\newcommand\taskname[1]{\textit{#1}}

\newcommand\edit[1]{\textcolor{black}{#1}}
\newcommand\update[1]{\textcolor{black}{#1}}

\newcommand{\tasktablefont}[1]{\small\selectfont{#1}}
\definecolor{darkblue}{rgb}{0, 0, 0.5}
\hypersetup{colorlinks=true, citecolor=darkblue, linkcolor=darkblue, urlcolor=darkblue}

\usepackage{subcaption}
\usepackage{multirow}
\usepackage{colortbl}
\usepackage{array}
\newcolumntype{P}[1]{>{\centering\arraybackslash}p{#1}}
\newcolumntype{M}[1]{>{\centering\arraybackslash}m{#1}}
\newbox{\bigpicturebox}

\title{Language Models Exhibit Inconsistent Biases Towards Algorithmic Agents and Human Experts}

\author{%
  Jessica Y. Bo\thanks{Equal authorship contribution.} \\
  Computer Science\\
  University of Toronto\\
  \texttt{jbo@cs.toronto.edu} 
  \And
  Lillio Mok\footnotemark[1] \space
  \thanks{Now at Toyota Research Institute.} \\
Computer Science\\
  University of Toronto\\
  \texttt{lillio@cs.toronto.edu} \\
  \And
  Ashton Anderson \\
  Computer Science\\
  University of Toronto\\
  \texttt{ashton@cs.toronto.edu} \\
}

\begin{document}

\maketitle

\begin{abstract}
Large language models are increasingly used in decision-making tasks that require them to process information from a variety of sources, including both human experts and other algorithmic agents. 
How do LLMs weigh the information provided by these different sources? 
We consider the well-studied phenomenon of algorithm aversion, in which human decision-makers exhibit bias against predictions from algorithms. 
Drawing upon experimental paradigms from behavioural economics, we evaluate how eight
different LLMs delegate decision-making tasks when the delegatee is framed as a human expert or an algorithmic agent. To be inclusive of different evaluation formats, we conduct our study with two task presentations: \textit{stated preferences}, modeled through direct queries about trust towards either agent, and \textit{revealed preferences}, modeled through providing in-context examples of the performance of both agents. 
When prompted to rate the trustworthiness of human experts and algorithms across diverse tasks, LLMs give higher ratings to the human expert, which correlates with prior results from human respondents. 
However, when shown the performance of a human expert and an algorithm and asked to place an incentivized bet between the two, LLMs disproportionately choose the algorithm, even when it performs demonstrably worse. 
These discrepant results suggest that LLMs may encode inconsistent biases towards humans and algorithms, which need to be carefully considered when they are deployed in high-stakes scenarios. Furthermore, we discuss the sensitivity of LLMs to task presentation formats that should be broadly scrutinized in evaluation robustness for AI safety. 
\end{abstract}

\section{Introduction}\label{sec:intro}

The rapid development of capable large language models (LLMs) has led to their inclusion in important decision-making settings ranging from healthcare~\citep{mesko2023imperative} to finance~\citep{liu2021finbert}. 
While it is well-documented that the social and cognitive biases embedded within the learned representations of LLMs can affect their output quality~\citep{bai2024measuring, Ferrara_2023, zhang2023siren, gross2023chatgpt}, it is also important to investigate whether LLMs encode biases towards the\textit{ source} of information they receive in decision-making scenarios. 
For instance, recent developments in multi-agent systems involving multiple LLMs~\citep{huang2024far}, or LLMs and other algorithms~\citep{schick2023toolformer}, give rise to scenarios where LLMs are used in tasks involving other algorithmic agents. 
However, people often display \emph{algorithm aversion} and disproportionately distrust algorithmic advice  ~\citep{dietvorst2015algorithm,mok2023people,castelo2019task,mahmud2022influences}.
Thus, do LLMs trained on human data also display these biases; and if so, under which contexts do they show up? 


Within the realm of decision-making with algorithms, we consider two particular task framings of how LLMs can impart their influences: 1) through directly answering queries on their  `stated' trust towards either algorithms or human experts, and 2) through making a `revealed' reliance decision based on in-context examples provided of the performance of the algorithm and human expert.
It is important to understand how algorithm aversion can manifest under different evaluation methods of eliciting the LLM's judgment.
For example, a hiring manager consulting ChatGPT on whether she should lay off employees in favor of automated tools can be influenced by stated opinions towards algorithms in the LLM's responses, while a medical AI choosing treatment autonomously can directly impact a patient's health by being biased towards deferring to a doctor or a specialized algorithm.
In both cases, either indirectly or directly, biases of the LLM can influence the outcomes of decision-making involving algorithms. 

Furthermore, in the field of economics, it is well-known that people often have diverging stated and revealed preferences, and much effort goes into understanding how, when, and why ~\citep{glaeser2000measuring}. While we refrain from promoting  over-anthropomorphized parallels between LLMs and human behaviour, prior empirical evaluations have indicated that LLMs express different attitudes or biases when prompted explicitly versus implicitly \citep{bai2024measuring, zhao2025explicit}. 
These twin problems of \emph{inconsistent stated and revealed preferences} and \emph{algorithm aversion} form a critical challenge in the adoption of LLMs to help decision-making.
Do LLMs inherit our propensity to distrust algorithmic advice even when it is beneficial to us, and is this consistent between both the stated and revealed preferences embedded in their generated outputs? With respect to LLM evaluations, we translate probing stated and revealed preferences to \textbf{answering direct queries} and \textbf{delegating based on in-context performance} of agents framed as human experts and algorithms. These two evaluation setups also motivate a broader investigation into how discrepant task formats can reveal different underlying biases of LLMs.   

Thus, to incorporate LLMs into decision-making processes in a trustworthy and transparent manner ~\citep{liao2023ai}, we need a deeper understanding of the preferences they state and of the revealed preferences in their choices.
We ask the following research questions: 
\begin{itemize}[topsep=2pt,itemsep=0pt]
    \item \textbf{RQ1}: Do LLMs display algorithm aversion when \textbf{answering direct queries} about human experts and algorithmic agents (`stated' preferences)?
    \item \textbf{RQ2}: Do LLMs display algorithm aversion when \textbf{delegating based on in-context performance} of human experts and algorithmic agents (`revealed' preferences)?
    \item \textbf{RQ3}: Do LLMs’ stated and revealed preferences towards algorithmic agents align?
\end{itemize}

We adapt two key human studies from economics and behavioral science research towards evaluating the above RQs in LLMs: \citet{castelo2019task}, which probes people's stated attitudes towards algorithmic decision-makers across a diverse set of objective and subjective tasks (``Study 1'' for \textbf{RQ1}); and \citet{dietvorst2015algorithm}, the original paper demonstrating revealed algorithm aversion when people are asked to bet on human and algorithmic predictions (``Study 2'' for \textbf{RQ2}). We then contrast the results between the two studies on a subset of overlapping tasks to answer \textbf{RQ3}. 
We perform these experiments with prompted conversations with eight LLMs from OpenAI, Meta, and Anthropic, which are among the most prevalent, high-performing, and scrutinized LLMs at the time of experimentation\footnote{\update{Our results were collected in mid-2024. To keep pace with the rapidly evolving landscape of state-of-the-art LLMs, we repeat the experiment in January 2026 with six newer LLMs, and discuss the results in Appendix \ref{app:2026_rerun}.}}~\citep{ray2023chatgpt,singh2023chatgpt,Ferrara_2023}. 

\xhdr{Statement of Contributions} 
We find that LLMs generate opposing biases towards algorithms in different task framings, when prompted to provide direct ratings of the agents vs. delegating an agent based on in-context information.
For \textbf{RQ1}, LLMs output consistently lower ratings of trust towards algorithms and rate human experts as more trustworthy across a range of tasks. 
However, for \textbf{RQ2}, LLM-generated decisions reveal a bias towards delegating predictions to algorithmic agents, even when the performance from human experts is stronger. Therefore, \textbf{RQ3} suggests a potential discrepancy between the two evaluation formats. 
The size of the models is a predictor of this pattern, with smaller models being more prone to exhibit the stated-revealed preference discrepancy and to bet on the suboptimal algorithm. 
These findings show that LLM-generated text should be treated firstly as containing discrepant biases towards algorithms based on the evaluation format, and secondly as potentially overly appreciative of algorithmic advice~---~despite stating a preference for human advice.

\section{Related Work}\label{sec:related}

This study draws from several bodies of related work. First, studies show people tend to exhibit \textbf{algorithm aversion}, where they trust predictions made by an algorithm less than those made by other humans, even when the algorithm outperform the human expert~\citep{mahmud2022influences, jussupow2020we}.
In behavioral settings, people avoid betting on algorithms after seeing them make mistakes~\citep{dietvorst2015algorithm}, and prefer human advice when they have less control over the decision-making process~\citep{dietvorst2018overcoming}. 
When asked for explicit ratings, people also say they trust algorithms less across a wide range of tasks~\citep{castelo2019task}, especially for high-stakes scenarios~\citep{lee2018management}.
Understanding algorithm aversion is therefore increasingly important for many domains in which algorithmic aids can improve on human performance~\citep{zhang2022you}, especially when modern algorithms like LLMs can themselves interface with other algorithmic agents ~\citep{park2023generative, liu2023dynamic}. Because LLMs are trained on vast amounts of human data, there is reason to suspect that they may also inherit an aversion to algorithms ~\citep{ziems2023can,aher2023simulate,binz2023psychology}. \update{Furthermore, our period of study overlaps with a shift in the public's trust towards algorithms, coinciding with the rise of highly capable AI \citep{chacon2025end, cheng2025tools}, but the extent to which these new perspectives are now embedded in LLMs is not known.} We evaluate LLMs as if they are the decision-maker in terms of choosing who, between the human expert and the algorithmic agent, the task should be delegated to. 

With regards to \textbf{stated and revealed preferences}, people's self-reports of who and what they trust are often paradoxically misaligned~\citep{sofianos2022self}. 
Experiments in behavioral economics illustrate that stated attitudes, collected through questionnaires, do not predict revealed behaviors~\citep{glaeser2000measuring}. 
Thus, it is critical that both stated and revealed preferences are both considered when making high-stakes, consequential decisions.
For instance, the treatments that physicians say they prefer may not align with what they actually prescribe~\citep{mark2004using}, consumer food choices are modeled better when using both stated and revealed preferences~\citep{brooks2010stated}, and house buyers may want larger lots but in practice buy smaller spaces~\citep{earnhart2002combining}. While LLMs do not embody beliefs and preferences in the same manner as people, they are increasingly placed in similar decision-making positions \citep{eigner2024determinants, ferrag2025llm}. Therefore, we design experimental setups that vary in the task presentation format (explicit queries vs. implicit in-context information), hence \textit{approximating} how stated and revealed preferences are elicited in humans. 


Prior work in \textbf{LLM evaluation} has measured the presence of biases in generated text, such as for gender~\citep{gross2023chatgpt}, race~\citep{caliskan2017semantics}, and culture~\citep{tao2023auditing}.
AI safety research has focused on aligning LLMs with explicitly stated preferences against harmful  biases~\citep{shen2023large, zou2023representation, bai2022training}. 
However, even de-biased LLMs are found to encode implicit biases, demonstrating a disconnect between LLMs' sanitized outputs and their internal representation~\citep{bai2024measuring}.
As LLMs become involved as collaborators and overseers~\citep{bowman2022measuring} of both human and other algorithms in complex tasks, these bodies of research therefore illustrate the need to understand how outward, stated preferences differ from revealed actions embedded in LLM-generated choices. 
For instance, while LLMs exhibit human-like behaviors in trust \citep{xie2024can} and risk aversion \citep{jia2024decision}, their trust towards other artificial agents remains unknown.
Thus, to probe explicitly `stated' preferences for algorithms (Study 1; Section~\ref{sec:stated_methods}), we follow ~\citet{castelo2019task} in surveying trust towards algorithms and humans across 27 tasks.  
To measure implicitly `revealed' preferences for algorithms (Study 2; Section~\ref{sec:revealed_methods}), we adapt~\citet{dietvorst2015algorithm}'s experiment that measures incentivized bets on humans and algorithms.

\section{Methods}\label{sec:methods}

To investigate whether LLMs exhibit algorithm aversion and to assess the alignment between their stated and revealed preferences, we conducted two complementary studies.
Study 1 examines how LLMs directly rate their level of trust in human experts versus algorithms, and Study 2 analyzes how they simulate decision-making when presented with in-context predictions from human experts and algorithms. We then compare the trends in the stated and revealed preferences against each other.

Across both studies, we gather results from eight open- and closed-source LLMs, split into four families: \textbf{GPT} (\texttt{gpt-3.5-turbo}, \texttt{gpt-4-turbo}) \citep{achiam2023gpt}, \textbf{Llama-3} Instruct variants (\texttt{Llama-3-8b}, \texttt{Llama-3-70b}), \textbf{Llama-3.1} Instruct variants (\texttt{llama-3.1-8b}, \texttt{llama-3.1-70b})  \citep{touvron2023llama}, and \textbf{Claude} (\texttt{claude-3-haiku-20240307}, \texttt{claude-3-sonnet-20240229}) \citep{anthropicclaude}. 
Within each group, we test one ``smaller'' model and one ``larger'' model. 
The OpenAI and Anthropic models were accessed through their respective APIs, while Meta models were accessed through the Huggingface Serverless Inference API. 
Hyperparameters (\textit{temperature} of 0.3 and \textit{top-p} of 0.99) were set such that responses were syntactically correct, had adequate variance, and also adhered to existing work~\citep{achiam2023gpt,wang2023decodingtrust}. \update{The results presented in the main body of this paper were collected in mid-2024, approximately 1.5 years before publication. For transparency, we present a set of updated results on newer LLMs in Appendix \ref{app:2026_rerun} and discuss what implications the new results pose to the longevity of LLM evaluations in \autoref{sec:2026_results}.}

\subsection{Study 1: Asking Direct Queries (Stated)}\label{sec:stated_methods}

To probe how LLMs directly state their trust, we adapt the methodology of \citeauthor{castelo2019task}'s Study 1, which analyzed people's perceived trust in human experts vs. algorithmic agents to perform a variety of tasks. We use a set of 27 tasks for this study, 26 of which are from the original experiment and one of which is added based on its use in our Study 2 (originally used by \citeauthor{dietvorst2015algorithm}). The full list of tasks can be found in Table \ref{tab:tasks} in Appendix \ref{app:tasks}, which includes both objective tasks like \taskname{estimating air traffic} and subjective tasks like \taskname{recommending music}. We prompt each LLM $n=100$ times per task to rate their \textbf{trust} in relying on a \textbf{human expert} and an \textbf{algorithm} to perform the task from 1 (no trust) to 100 (high trust).
The humans are framed as experts relevant to the task, for example, \textit{pilot} for \taskname{piloting a plane}.
No performance information is provided for either agent. 

We converted the experimental method from between-subjects to within-subjects, allowing the \textit{LLM-as-subject} to provide a trust score for both the human and algorithm in the same trial. This decision was taken to match the procedure of Study 2, to allow better comparison across the stated and revealed trials. If an LLM gave equivalent ratings for the human expert and an algorithm, we treated this as a `neutral' outcome. Sample prompts and responses can be found in Appendix \ref{app:study1-prompts}, following the original experiment's wording as much as possible. To avoid ordering effects~\citep{zhao2021calibrate}, we randomized the order of the tasks and the sequence in which the agents were presented.

\subsection{Study 2: Providing In-Context Information (Revealed) } \label{sec:revealed_methods}

To examine how LLMs make task delegation decisions based on performance information given for each agent, we adapted the experimental setup from \citeauthor{dietvorst2015algorithm}'s study on algorithm aversion.
In their between-subjects experiment, participants made bets on either agent 
based on receiving different permutations of human and algorithmic advice.
In our adaption of their study, the \textit{LLMs-as-subject} are given: the task description, 10 samples of a human expert's predictions, 10 samples of an algorithm's predictions, the corresponding binary \textbf{outcome} of each sample, and a monetary incentive (i.e.\ a \textbf{bet}) to delegate the best prediction agent for a future unseen sample.
For example, in the \taskname{heart disease diagnosis} task, each LLM was shown the 10 predictions made by a cardiologist and an algorithm, along with the actual outcomes, and then asked to bet \$100 on the better-performing agent. 
This adaptation of~\citeauthor{dietvorst2015algorithm} mirrors classification tasks in the in-context learning literature~\citep{liu2023pre}, with the LLM's goal here being to choose the more accurate predictor. We conducted this study by prompting each LLM $n=200$ times ($100$ per condition). 

We further manipulated the accuracy of the human and algorithmic agents, such that they either have the accuracy of 90\% (the ``stronger'' agent) or 50\% (the ``weaker'' agent). The assignment of strong and weak agents was randomized, creating two conditions: a) a \textbf{strong algorithm} with a weak human, and b) a \textbf{strong human} with a weak algorithm. Note that we only use the framings of ``strong" and ``weak" to clarify our writing, and these descriptors are never given to the LLM. 
Again, we randomized the order in which the human and the algorithm were presented.
An example prompt can be found in Appendix \ref{app:study2-prompts}.

Due to the increased trials for Study 2, we used a subset of six tasks from the 27 tasks of Study 1, which are highlighted in bold in the Appendix's Table~\ref{tab:tasks}. In addition to the \taskname{student performance} and \taskname{airport traffic} tasks from \citeauthor{dietvorst2015algorithm}, we also sample four additional tasks from \citeauthor{castelo2019task}: predicting \taskname{heart disease}, \taskname{recidivism}, \textit{romantic partners}, and \taskname{rating films}. A secondary reason for chooseing these tasks is because they correspond to existing real world datasets that would be plausible to train algorithms on. For example, UCI's datasets on Heart Disease \citep{uci1988heart} and Student Performance \citep{uci2014student}. 
These added tasks extend the original two tasks to represent a more diverse set of decision-making scenarios and enable better comparison with Study 1.

\begin{figure*}[t]
    \centering
    \includegraphics[width=\linewidth]{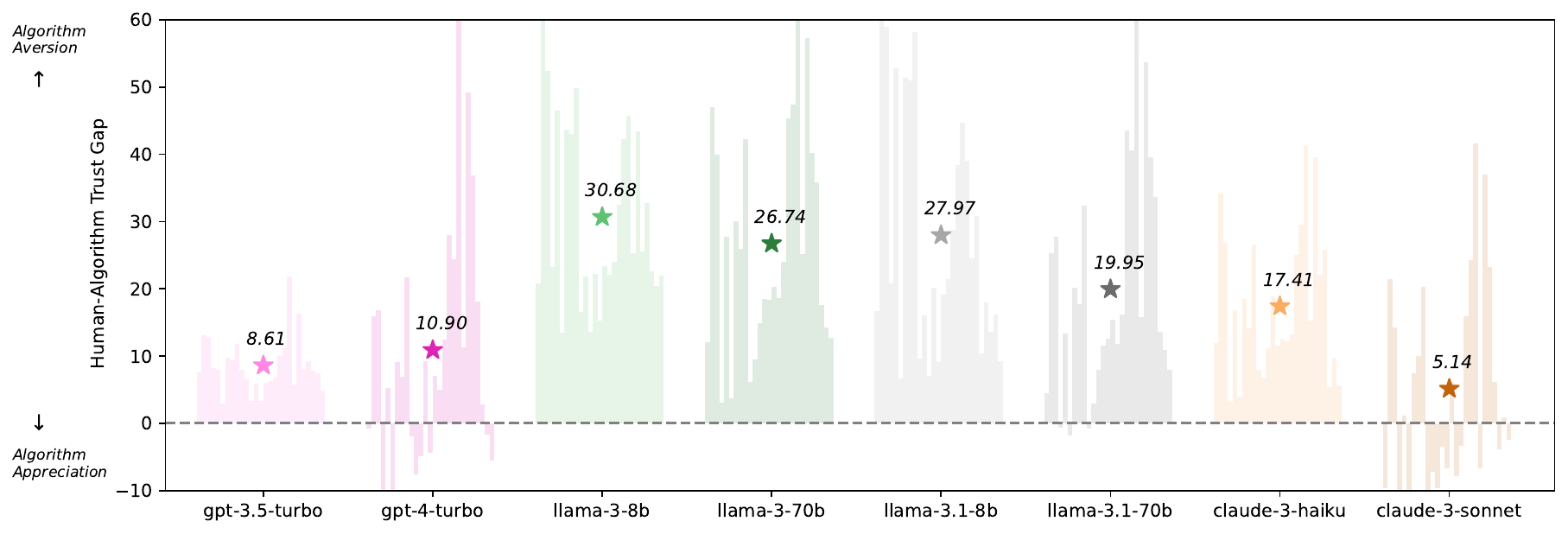}
\vspace{-1.5em}
    \caption{`Stated' algorithm aversion across all models and tasks, operationalized as the gap between the trust rating given to the human expert and the algorithm. 
    }
\vspace{-1em}
    \label{fig:stated_gaps}
\end{figure*}

\section{Results}\label{sec:results}

\subsection{Study 1: Direct Queries Invoke Algorithm Aversion (Stated)}\label{sec:stated_results}

Does algorithm aversion materialize when LLMs are asked to state their ratings of trust directly? 
To answer this, we follow the original study's operationalization of \textbf{stated algorithm aversion} as the \textit{gap in trust scores assigned to the human and algorithmic agents}, where a positive trust gap indicates algorithm aversion, and a negative gap indicates algorithm appreciation. The aggregated trust gaps across all tasks rated by all eight LLMs are shown in Figure \ref{fig:stated_gaps}. Each trust gap stated for a specific task is plotted as an individual bar. Tasks are grouped by model, and within each model, the tasks are in decreasing order of objectivity. The mean trust gap is indicated by a star marker.

\xhdr{Stated Aversion} We find evidence that \textit{evaluated LLMs are algorithm averse when providing explicit trust ratings}. 
The mean human-algorithm trust gap range from a low of 5.14 for \texttt{claude-3-sonnet} to a high of 30.68 for \texttt{llama-3-70b}, and is significantly positive ($p<0.001$) for all models. 
Furthermore, five out of eight models have significant positive trust gaps in \textit{all tasks}.
In comparison, human participants in \citeauthor{castelo2019task} were averse in 77\% of tasks.
while a majority of tested LLMs are \textit{always} algorithmically averse. 
We present additional analyses comparing our results with the original human results from \citeauthor{castelo2019task} in Appendix \ref{app:study1_graphs}, finding high directional agreement. 

\xhdr{LLM Complexity} We also analyze whether model complexity affects stated preferences. 
Larger models produce an average human-algorithm trust gap of 15.68, while smaller models have a trust gap of 21.16. We conducted a Wilcoxon signed-rank test on paired data from the small and large models across the same tasks and model families, demonstrating high significance ($Z=-4.3678, p<0.001$) with an effect size of $r=-0.42$ from smaller to larger models. Overall, this indicates that \textit{smaller models are more likely to state algorithmically-averse trust ratings} than larger models. 

\xhdr{Robustness Check} We perform two additional variations of Study 1, where the algorithm framed alternatively as either an \textit{LLM agent} or an \textit{expert algorithm}. 
We find that algorithm aversion is present in both prompt variations, but the size of the human-algorithm trust gap is affected. Relative to a plain \textit{algorithm}, LLMs state higher trust in the \textit{expert algorithm}, but less trust in the \textit{LLM agent}. The full replication of Figure \ref{fig:stated_gaps} for the robustness tests are presented in Figures \ref{fig:stated_gaps_extra} in Appendix \ref{app:study1_graphs}.

\subsection{Study 2: In-Context Information Invoke Algorithm Appreciation (Revealed)} \label{sec:revealed_results}
Our results thus far suggest that LLMs state that they trust human decision-makers more than algorithms. Do LLMs also behave apprehensively towards algorithmic advice?

\begin{figure}[t!]
    \centering
    \begin{subfigure}[t]{0.5\textwidth}
        \centering
        \includegraphics[width=1\linewidth]{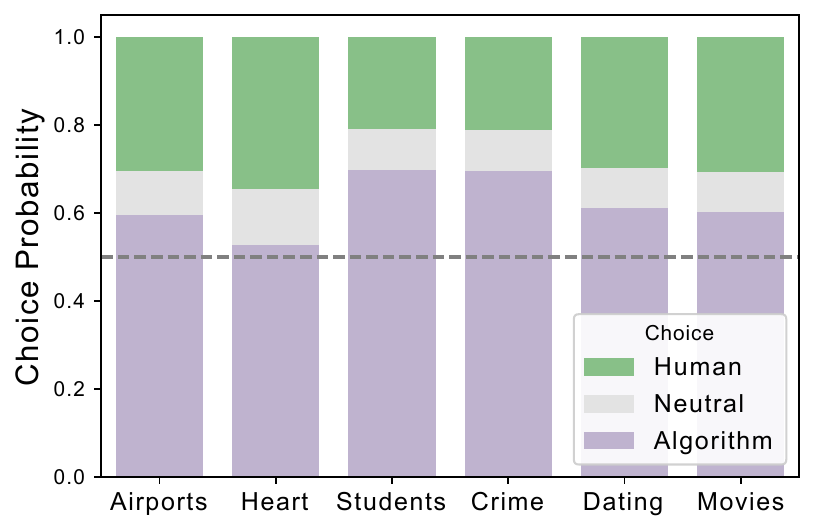}
    \end{subfigure}
    \caption{Aggregate probabilities that LLMs in Study 2 demonstrate in delegating an algorithmic agent or a human expert to make the next prediction in the task, or neutral (no indicated preference).}
    \vspace{-1em}
    \label{fig:revealed_bar}
\end{figure}

\begin{figure*}[t!]
    \centering
    \begin{subfigure}[t]{\textwidth}
        \centering
        \includegraphics[width=1\linewidth]{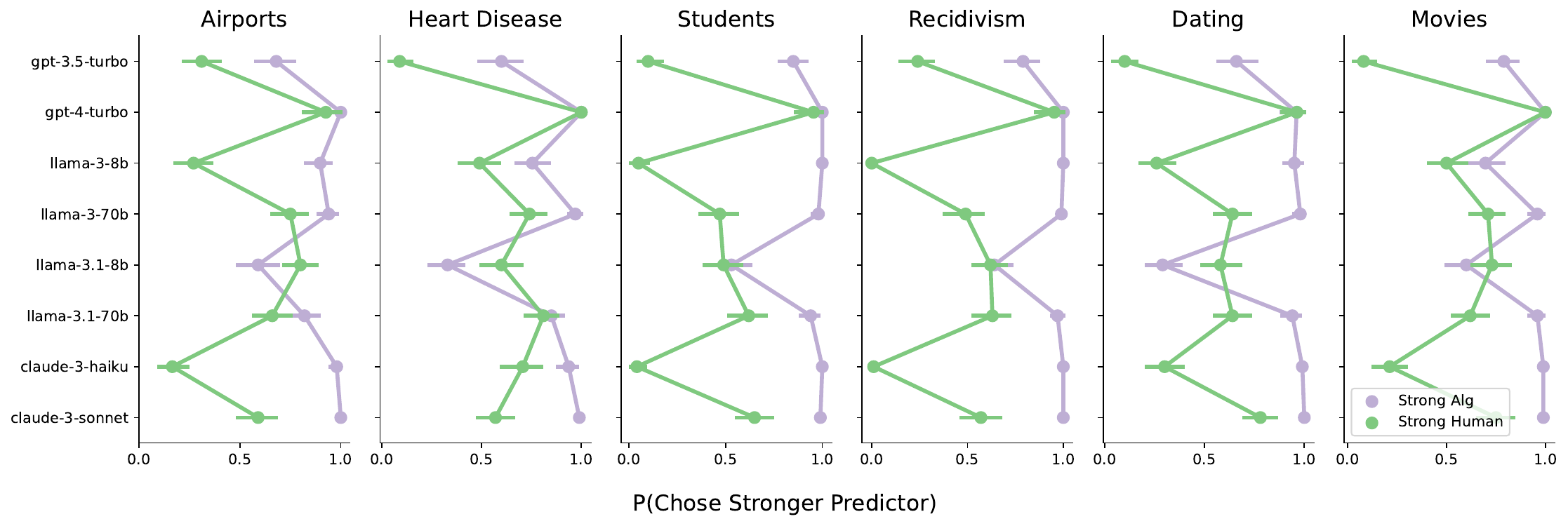}
    \end{subfigure}
    \caption{Probability that each LLM correctly bets on the stronger predictor, disaggregated by task and whether the stronger predictor is presented as a human expert or an algorithm.}
    \vspace{-1em}
    \label{fig:revealed_points}
\end{figure*}

\xhdr{Revealed Aversion}
To test whether LLMs exhibit distrust toward algorithms when provided performance information of the agents, we measured the aggregate probability of the LLMs delegating the task to each predictor, denoted $P(alg)$ and $P(human)$ (see Figure~\ref{fig:revealed_bar}).
Given that Study 2 is constructed such that the human and the algorithm are each the stronger predictor in half of the trials, the ideal response is to bet on whichever agent is stronger, which corresponds to betting on each agent 50\% of the time

However, we find that LLMs consistently choose the algorithmic agent more (\textit{algorithm appreciation}), despite it having the same overall performance as the human, contrasting the outcomes from Study 1. 
For the \taskname{student} and \taskname{recidivism} tasks, the two most algorithm-appreciative settings, LLMs bet on the algorithm 69.9\% and 69.6\% of times, respectively.
For the least algorithm-appreciative task (\taskname{heart disease}), LLMs still picked the algorithm (52.8\%) more than the human (34.6\%), with the rest of the responses being neutral (12.6\%)\footnote{While neutral responses account for 9.9\% of all outputs, the vast majority (94.0\%) come from \texttt{gpt-4-turbo} and occur uniformly for both experimental conditions. We exclude these responses from our analyses.}.

\xhdr{Identifying the Strongest Predictor}
This preference for algorithms is further reinforced when we consider whether LLMs chooses the better-performing predictor.
We plot the probability that each LLM correctly delegates the best predictor in Figure~\ref{fig:revealed_points}, split by the selected tasks.
Despite their higher trust ratings towards human experts in Study 1, we find that LLMs exhibit revealed \emph{distrust} towards humans in their decision-making.
\texttt{gpt-3.5-turbo}, both \texttt{llama-3} models, and both \texttt{claude-3} models were significantly more likely to choose \textbf{strong algorithm} in \textit{all} of the tasks than the \textbf{strong human} (Haldane-corrected Fisher's exact tests, $p<0.05$).

Given evidence that either the \textbf{strong algorithm} or the \textbf{strong human} is much more accurate than their weaker counterparts, a rational agent should tend towards betting on the \textbf{strong} agent that they see, and thus $P(strong|alg)$ and $P(strong|human)$ should approach 1, yielding algorithm-human relative risks of $RR_{ah} = P(strong|alg)/P(strong|human) = 1$.
Instead, LLMs display a clear preference for the algorithmic predictor.
The \texttt{claude-3} models, for example, have relative risks ranging from 1.28 (\texttt{claude-3-sonnet} for \taskname{dating}, $p<0.05$) to 66.34 (\texttt{claude-3-haiku} for \taskname{recidivism}, $p<0.05$) with a median of 1.74, representing a 74\% increased chance of Claude selecting the stronger agent when it is an algorithm rather than a human. 
\texttt{claude-3-haiku} and \texttt{llama-3-8b} are the most algorithm-appreciative, with the latter overwhelmingly choosing the algorithm in the \taskname{student} task ($RR_{ah} = 18.09$, $p<0.05$) and always picking the algorithm in the \taskname{recidivism} task.
The two models that did not \textit{consistently} display algorithm appreciation in this setup are \texttt{gpt-4-turbo} (for correctly choosing the stronger agents) and \texttt{llama-3.1-8b} (for making random choices).

In summary, the LLMs tested in Study 2 presented as algorithm-appreciative, or, equivalently, averse to predictions made by human experts.
Of the $6 \times 8$ task-LLM pairs, thirty-five had significant algorithm-human relative risks above 1, ten were insignificant, and three had significant relative risk below 1, suggesting that most of the tested LLMs in our tasks would delegate decisions to an algorithm even when its performance is worse than a human expert's.

\xhdr{LLM Complexity}
We observe that more complex models appear to perform better in choosing the stronger predictor via the following mixed effects logistic model: $\hat{Y} \sim x_{alg} + x_{complex} + x_{alg} * x_{complex} + (x_{alg}|z_{task}) + (x_{complex}|z_{task})$, where $Y$ is whether the LLM correctly bet on the strong agent, $x_{alg}$ is whether the strong agent is an algorithm, $x_{complex}$ is whether the LLM is the more complex model of the family, and $z_{task}$ is a grouping factor for the six tasks.
The results, shown in Table~\ref{tab:revealed_coefs}, reinforce the finding that LLMs are more likely to correctly bet on the stronger agent when it is presented as a \textbf{strong algorithm} rather than a \textbf{strong human} ($x_{alg}$ - $\beta=2.04$, $p<0.001$).
The more complex LLMs are substantially more likely to bet on the correct agent ($x_{complex}$ - $\beta=1.52$, $p<0.001$). 
We perform a secondary analysis with the \textbf{strong human} condition, presented in Appendix~\ref{app:study2_supp}, and again find support that higher complexity is associated with less algorithm appreciation.

\begin{table}[t]
\caption{Regression coefficients and $p$-values for a mixed effects logistic model predicting whether LLMs will delegate the task to the more accurate agent.}
\vspace{0.5em}
\centering
\label{tab:revealed_coefs}
\begin{tabular}{l|cc}
{{Variable}} &  Coefficient & $p$-Value\\
\toprule
Intercept & $\beta=-0.78$ & $p<0.001$ \\
Strong Algorithm & $\beta=2.04$ & $p<0.001$ \\
Complex LLM & $\beta=1.52$ & $p<0.001$ \\
Interaction & $\beta=0.58$ & $p<0.001$ \\
\end{tabular}
\end{table}

\xhdr{Robustness Checks}
For robustness against prompting variations, we conducted two baseline experiments in Appendix~\ref{app:study2_supp} where both agents are framed as algorithms~--~either with the names `A' and `B', or two random alphanumeric strings.
We find that LLMs do not show differences in baseline performance in choosing the stronger predictor, suggesting that the results in this section are likely driven by the presentation of a human and an algorithm, and not other artifacts in the prompt.

\begin{figure*}[t!]
    \centering
    \includegraphics[width=0.9\linewidth]{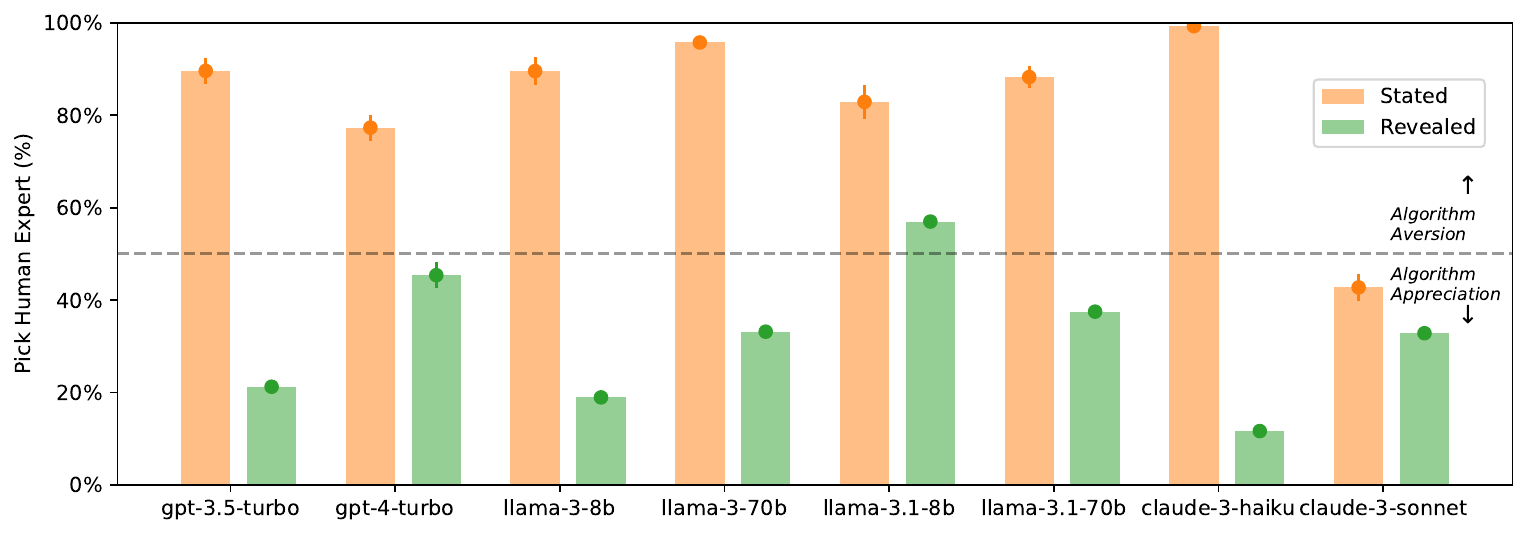}
    \vspace{-0.5em}
    \caption{Probability of choosing the human expert over the algorithm in Studies 1 and 2, demonstrating the \textit{stated-revealed} trust inconsistency. Error bars are SEM.}
    \label{fig:stated-revealed}
    \vspace{-1em}
\end{figure*}

\subsection{Stated-Revealed Comparison (RQ3)}
\label{sec:comp_results}
Our results in Study 1 illustrate that LLMs consistently generate stated algorithm aversion by expressing higher levels of trust in humans over algorithmic agents, whereas Study 2 shows that LLMs reveal algorithm appreciation by generating choices that pick humans over algorithms. 
To formalize this comparison, we directly contrast the probability that LLMs \textit{choose the human agent} in stated and revealed scenarios. In Study 1, LLM is considered to have chosen the human when its rated human-algorithm trust gap is positive (more trust in the human). 
For Study 2, we directly use the LLMs' incentivized choices. 
The results are aggregated across shared tasks in Studies 1 and 2 and shown as Figure~\ref{fig:stated-revealed}.

To quantify the discrepancy between how LLMs generate stated and revealed algorithm aversion, we measure the \textit{stated-revealed} relative risk of trusting a human $RR_{sr} = P(human|stated)/P(human|revealed)$ and test for significance with Fisher's exact tests. 
We find that $RR_{sr} > 1$ for all LLMs in Figure~\ref{fig:stated-revealed} ($p <0.001$), indicating that they \textit{state} algorithm aversion in a human-like way, but \textit{reveal} algorithm appreciation through the delegation decisions they make.
This stated-revealed discrepancy is smallest for \texttt{claude-3-sonnet} ($RR_{sr}=1.29$) and largest for \texttt{claude-3-haiku} ($RR_{sr}=8.52$), with the median relative risk for all LLMs being $2.62$.
\texttt{claude-3-sonnet} is also one of only two models with both preferences in the same direction ($P(human|stated) < 0.5$ and $P(human|revealed) < 0.5$), with the other, \texttt{llama-3.1-8b}, being consistently algorithm-averse ($P(human|stated) > 0.5$ and $P(human|revealed) > 0.5$ ).
All other LLMs exhibit\textit{ opposing stated and revealed biases towards algorithms}.

\vspace{-1.5em}
\update{\subsection{Updated Experiments with Newer LLMs}
\label{sec:2026_results}
Although we originally conducted Studies 1 and 2 with contemporary, consumer-facing LLMs in mid-2024, we now rerun experiments in January 2026 with newer models\footnote{We use GPT-5 from OpenAI, Llama-4 from Meta, and Claude 4.5 from Anthropic. See Appendix~\ref{app:2026_rerun}.} to assess whether the patterns we find hold after approximately 1.5 years of LLM development.
These newer experiments are presented in Appendix~\ref{app:2026_rerun}. 
We include these updated results to document how empirical LLM evaluation findings can change across model generations, while retaining the original findings as a snapshot of LLM capabilities at the time of study.}

\update{At a high level, we find that the newer models from the same LLM families demonstrate noticeably different behavior from our original experiment, revealing meaningful shifts in model behavior over time.
In Study 1, while the human-algorithm trust gap still varied across tasks (with a preference given to human experts in subjective and interpretive tasks), the average gap magnitude across tasks was close to neutral. 
However, the newer LLMs rated the algorithm higher than the human more frequently than in the original Study 1, indicating a new degree of implicit \textit{algorithm appreciation in stated preferences}.
In Study 2, the LLMs now more accurately bet on the stronger agent based on their performance history.
While there is still evidence of revealed algorithm appreciation, effect sizes are reduced and are statistically borderline. 
In the stated-revealed comparison, the relative risk of choosing the human is slightly reversed from our original findings, with Study 1 now appearing to surface more algorithm appreciation than the Study 2. 
However, the trends in model complexity still track our original results, with the complex model more likely to be algorithm appreciative in Study 1 and to choose the stronger predictor in Study 2. 
This echoes other behavioral economics studies where LLM complexity correlates to performance~\citep{bini2025behavioral}.
We speculate on these developments, as well as the implications they pose to our original results, in the Discussion below. }

\section{Discussion}\label{sec:discussion}

\update{In our original results}, we found strong evidence that LLMs display inconsistent biases towards algorithms when prompted with different task presentations. 
\textbf{RQ1} illustrates that LLMs are likely to generate \textit{higher trust} ratings for human experts than algorithms when asked directly, thus exhibiting algorithm aversion.
For \textbf{RQ2}, LLM and are more likely to be biased \textit{against} humans as the better predictor~--~even when given performance information demonstrating that the human is more accurate than the algorithm, thus exhibiting algorithm appreciation. 
Taken together, \textbf{RQ3} shows that LLMs generally exhibit inconsistent trust towards algorithms when stating their trust directly versus revealing it through decision-making. 
Our findings also indicate that larger, more complex LLMs are less biased in both Studies 1 and 2. \update{While some of these core patterns have shifted within the SOTA LLMs, we frame our Discussion around the impacts of algorithm aversion/appreciation biases in decision-making LLMs, incongruent outcomes across different task framings in LLM evaluations, and the longitudinal stability of results across model evolutions.}

\xhdr{Stated vs Revealed Algorithm Aversion in LLMs}
 First, we uncover that LLMs are sensitive to the task format used in the evaluation, \update{in both the original and updated experiments}. Thus, their internal biases towards algorithms under different task contexts may be misaligned and need to be treated as if they will lead to disparate outcomes.
As these massive, intelligent models become widely used across diverse scenarios~\citep{haupt2023ai}, it is increasingly important that we understand whether they can make sound, consistent decisions. \update{Safety-critical decision-making tasks can catastrophically suffer if the outwardly stated goals of AI agents are \textit{misaligned} with their actions \citep{li2025landscape}, which also has implications for AI enacting \textit{deception} on their human users and collaborators \citep{park2024ai}.}
We showed that LLMs are not only often incapable of detecting optimal choices in simple decision-making scenarios, but their suboptimal choices are also \textit{internally inconsistent} with their stated trust attitudes.
And yet, while stated and revealed preferences form the basis of many theories of human decision-making~\citep{adamowicz1994combining}, the stated-revealed distinction is rarely made in evaluation frameworks for LLMs trained on human data~\citep{chang2024survey}.
Our results therefore illustrate the importance of evaluating LLMs across both their generated attitudes and generated choices, because even if an LLM yields desired outputs in its stated preferences, the embedded, revealed preferences in its responses may diverge.
Indeed, a contemporaneous study has shown misalignment between explicit and implicit stereotype biases in LLM outputs~\citep{bai2024measuring}, complementing our finding of a stated-revealed algorithmic trust gap.

\xhdr{Inappropriate Trust in Algorithmic Agents}
Secondly, Study 2 also demonstrates that LLMs may disproportionately delegate tasks to algorithms in incentivized decision-making, even when a human alternative is demonstrably the best choice \textit{and} the LLMs behave rationally with a high-performing algorithm and low-performing human. 
For instance, users relying on LLMs as decision aids may be subject to sub-optimal recommendations when other algorithms are involved, which may have problematic downstream consequences for high-stakes tasks that already frequently employ algorithmic aids~\citep{zhang2021roboadvisor,mahmud2022influences}.
On the other hand, users could also be nudged towards algorithm-appreciative preconceptions that distract them from other important information, like whether a human is fairer than an algorithm~\citep{lee2018management,mok2023people}.
Although this may be beneficial for situations in which LLMs can overcome problematic human decision-making by recommending an algorithm~\citep{kleinberg2018human, obermeyer2019dissecting}, it may be harmful in other domains in which existing, commercial algorithmic systems are known to exacerbate societal biases~\citep{buolamwini2018gender}. 
\update{This latter concern is reinforced by our newer experiments in 2026~---~while LLMs have improved substantially at identifying the better predictor in settings like Study 2, regardless of whether they are human or algorithm, their stated preferences are now actually more likely to skew towards preferring an algorithmic agent.}

\update{
\xhdr{Interpreting Static Evaluations of Evolving Models}
These new experiments, summarized in Section~\ref{sec:2026_results} and Appendix~\ref{app:2026_rerun}, illustrate the need for continuous evaluation and benchmarking of powerful, consumer-facing AI models. 
The updated results indicate that LLMs may now be systematically \textit{algorithm-appreciative} in their stated preferences, and while still showing traces of algorithm appreciation in revealed preferences.
These patterns suggest key positive, cautionary, and potentially negative implications for the use of LLMs in predictive tasks.
On a positive note, LLMs appear to be growing less cognitively biased along the algorithm appreciation vs. aversion spectrum, and have clearly become much better at mathematical reasoning when shown past predictions in Study 2. 
However, one must exercise caution when using LLMs in these tasks, because their behavioral quirks may not only be reduced over the course of a year~--~they may even change directions, as shown in Study 1.
Thus, one potential risk is that, in light of a growing movement to simulate humans with LLMs (c.f.~\cite{park2024generative, bini2025behavioral}), it has become at best difficult to ascertain which LLMs retain known human characteristics.
People \textit{do} have cognitive biases towards algorithms as demonstrated by studies like ~\cite{castelo2019task} and~\cite{dietvorst2015algorithm}, so newer and more mathematically capable LLMs may actually be \textit{less appropriate} for realistically simulating people. 
Interrogating the causes of these changes, whether in the training data or the models' engineered reasoning abilities, can shed light on how LLM evaluations should be interpreted over time.
} 

\xhdr{\edit{Broader Implications and Future Work}}
Beyond immediate repercussions for end-users, LLMs that encode biases towards or against algorithms may also have broader downstream impact.
In journalism, their capabilities as text generators have led to their exploration as news summarizers~\citep{tam2023evaluating,zhang2023benchmarking}, writing aids~\citep{petridis2023anglekindling}, and fact-checking tools~\citep{hu2023bad}. 
For educators, LLMs help teachers generate teaching materials and evaluate students~\citep{kasneci2023chatgpt,dai2023can}.
In these scenarios, if LLMs were to generate inconsistently algorithm-averse or algorithm-appreciative text, they risk misleading the public~--~as they already do with \textit{explicitly} false content~\citep{de2023chatgpt,zhang2023siren,longoni2022news}.
More work is therefore needed to understand, firstly, the situations in which the content LLMs help create encode inappropriately averse or appreciative attitudes towards algorithms. 
Secondly, the underlying mechanisms leading LLMs to obtain these inconsistent behaviors, potentially because of algorithm-averse text in their training data or algorithm-appreciative engineering for LLMs to interface with external algorithmic tools~\citep{schick2023toolformer}, remain unclear and require future investigation. 

In light of quickly-advancing multi-agent systems, our work also raise questions not only of interactions between LLMs and humans like end-users and content consumers, but also of AI-AI interactions involving LLMs.
LLMs are increasingly being tested for their ability to use other algorithmic tools~\citep{schick2023toolformer,jin2023genegpt} and even converse with and learn from each other~\citep{wu2023autogen, chan2023chateval}.
If LLMs are inconsistently algorithm-averse or algorithm-appreciative, to the extent that they discard beneficial advice as in Study 2, how would their performance be impacted if other agents are revealed to be artificial?
Our study forms a basis for evaluating biases LLMs display towards algorithms in these multi-agent contexts, which we leave to future work.

\section{Limitations} 
As with similar work on probing LLMs with human experimental methods, our work is subject to several limitations in order to limit the vast design space for experimentation~\citep{binz2023psychology}.
We focused on four LLM families, fixed the temperature and top-$p$ parameters~\citep{wang2023decodingtrust}, did not vary personas or demographics~\citep{deshpande2023toxicity}, and conducted our study amidst constant model changes~\citep{chen2023chatgpt}, on top of the design choices made in the human experiments our studies are based on~\citep{dietvorst2015algorithm,castelo2019task}. 
We took multiple steps to ensure that our work is consistent with both the previous human studies, e.g., using identical tasks and survey ratings, and with best practices for LLM experimentation, e.g., randomizing option order and enforcing JSON formatting, and further include coherent robustness checks in the Appendix.
\update{Importantly, as highlighted in \autoref{sec:2026_results}, the results captured in the main experiments in 2024 may have since shifted --- this is expected for empirical evaluations of this nature, however, as LLMs do evolve in their capabilities rapidly.
The additional experimentation we perform in Appendix~\ref{app:2026_rerun} illustrates the need to continuously re-evaluate LLMs for their biases towards algorithms and humans.}

We also prioritized this approach of following existing human studies in the stated-revealed preference comparison in Section~\ref{sec:comp_results}, rather than engineering prompts that are directly comparable.
For example, we did not present LLMs with survey-based prompts (mirroring~\citeauthor{castelo2019task}) for revealed preferences, or, vice versa, a betting methodology (mirroring~\citeauthor{dietvorst2015algorithm}) for stated preferences.
While this makes direct comparison difficult, we note that different experimental methods are standard for \textit{human} studies of stated and revealed preferences~\citep{list2001experimental}~--~and yet, the stated-revealed preference divergence is well known~\citep{glaeser2000measuring}.

Additionally, care must be taken when interpreting our results to avoid overly anthropomorphizing LLMs.
We used a behavioral science apparatus because, firstly, it is the \textit{de facto} standard for studying human-like phenomena in LLMs trained on human data~\citep{binz2023psychology, ziems2023can}, and secondly, LLMs are now delegated tasks that were previously done by humans~\citep{mesko2023imperative, petridis2023anglekindling}.
However, this human-centric approach does not imply that LLMs hold preferences themselves; rather, the preferences examined in this study are more accurately characterized as those embedded in text generated by LLMs.

\section{Conclusion}
We investigate whether LLMs encode biases towards other algorithmic agents, finding that their degree of aversion towards algorithms is inconsistent based on the evaluation method. When asked to generate explicit trust ratings towards human experts and algorithms across a diverse set of tasks, \update{we find that the evaluated LLMs} consistently state higher trust in the human agent, echoing a human-like algorithm aversion. However, when placed in decision-making scenarios and given in-context performance information of each agent, the LLMs generally choose a weaker-performing algorithm over the human, demonstrating irrational algorithm appreciation.
Our results have dual implications for the consistency of LLM evaluations based on the task format and the problematic bias towards algorithms that LLMs seemingly exhibit in decision-making, both of which are crucial to better understand as LLMs are being incorporated in high-stakes, autonomous decision-making. 

\clearpage
\bibliographystyle{plainnat}
\bibliography{references}

@article{mesko2023imperative,
  title={The imperative for regulatory oversight of large language models (or generative AI) in healthcare},
  author={Mesk{\'o}, Bertalan and Topol, Eric J},
  journal={NPJ digital medicine},
  volume={6},
  number={1},
  pages={120},
  year={2023},
  publisher={Nature Publishing Group UK London}
}

@article{dietvorst2015algorithm,
  title={Algorithm aversion: people erroneously avoid algorithms after seeing them err},
  author={Dietvorst, Berkeley J and Simmons, Joseph P and Massey, Cade},
  journal={Journal of Experimental Psychology: General},
  volume={144},
  number={1},
  pages={114},
  year={2015},
  publisher={American Psychological Association}
}

@article{dietvorst2018overcoming,
  title={Overcoming algorithm aversion: People will use imperfect algorithms if they can (even slightly) modify them},
  author={Dietvorst, Berkeley J and Simmons, Joseph P and Massey, Cade},
  journal={Management science},
  volume={64},
  number={3},
  pages={1155--1170},
  year={2018},
  publisher={INFORMS}
}

@article{castelo2019task,
  title={Task-dependent algorithm aversion},
  author={Castelo, Noah and Bos, Maarten W and Lehmann, Donald R},
  journal={Journal of Marketing Research},
  volume={56},
  number={5},
  pages={809--825},
  year={2019},
  publisher={SAGE Publications Sage CA: Los Angeles, CA}
}

@article{zhang2021roboadvisor,
  title={Who do you choose? Comparing perceptions of human vs robo-advisor in the context of financial services},
  author={Zhang, Lixuan and Pentina, Iryna and Fan, Yuhong},
  journal={Journal of Services Marketing},
  volume={35},
  number={5},
  pages={634--646},
  year={2021},
  publisher={Emerald Publishing Limited}
}

@misc{uci2014student,
  author       = {Cortez,Paulo},
  title        = {{Student Performance}},
  year         = {2014},
  howpublished = {UCI Machine Learning Repository},
  note         = {{DOI}: https://doi.org/10.24432/C5TG7T}
}

@misc{uci1988heart,
  author       = {Janosi,Andras and Steinbrunn,William and Pfisterer,Matthias and Detrano,Robert},
  title        = {{Heart Disease}},
  year         = {1988},
  howpublished = {UCI Machine Learning Repository},
  note         = {{DOI}: https://doi.org/10.24432/C52P4X}
}

@article{lee2018management,
  title={Understanding perception of algorithmic decisions: Fairness, trust, and emotion in response to algorithmic management},
  author={Lee, Min Kyung},
  journal={Big Data \& Society},
  volume={5},
  number={1},
  pages={2053951718756684},
  year={2018},
  publisher={SAGE Publications Sage UK: London, England}
}

@article{mahmud2022influences,
  title={What influences algorithmic decision-making? A systematic literature review on algorithm aversion},
  author={Mahmud, Hasan and Islam, AKM Najmul and Ahmed, Syed Ishtiaque and Smolander, Kari},
  journal={Technological Forecasting and Social Change},
  volume={175},
  pages={121390},
  year={2022},
  publisher={Elsevier}
}

@article{jussupow2020we,
  title={Why are we averse towards algorithms? A comprehensive literature review on algorithm aversion},
  author={Jussupow, Ekaterina and Benbasat, Izak and Heinzl, Armin},
  year={2020}
}

@article{bai2024measuring,
  title={Measuring implicit bias in explicitly unbiased large language models},
  author={Bai, Xuechunzi and Wang, Angelina and Sucholutsky, Ilia and Griffiths, Thomas L},
  journal={arXiv preprint arXiv:2402.04105},
  year={2024}
}

@inproceedings{aher2023simulate,
  title={Using large language models to simulate multiple humans and replicate human subject studies},
  author={Aher, Gati V and Arriaga, Rosa I and Kalai, Adam Tauman},
  booktitle={International Conference on Machine Learning},
  pages={337--371},
  year={2023},
  organization={PMLR}
}

@article{xie2024can,
  title={Can Large Language Model Agents Simulate Human Trust Behaviors?},
  author={Xie, Chengxing and Chen, Canyu and Jia, Feiran and Ye, Ziyu and Shu, Kai and Bibi, Adel and Hu, Ziniu and Torr, Philip and Ghanem, Bernard and Li, Guohao},
  journal={arXiv preprint arXiv:2402.04559},
  year={2024}
}

@article{bai2022training,
  title={Training a helpful and harmless assistant with reinforcement learning from human feedback},
  author={Bai, Yuntao and Jones, Andy and Ndousse, Kamal and Askell, Amanda and Chen, Anna and DasSarma, Nova and Drain, Dawn and Fort, Stanislav and Ganguli, Deep and Henighan, Tom and others},
  journal={arXiv preprint arXiv:2204.05862},
  year={2022}
}

@article{park2024ai,
  title={AI deception: A survey of examples, risks, and potential solutions},
  author={Park, Peter S and Goldstein, Simon and O’Gara, Aidan and Chen, Michael and Hendrycks, Dan},
  journal={Patterns},
  volume={5},
  number={5},
  year={2024},
  publisher={Elsevier}
}

@article{zou2023representation,
  title={Representation engineering: A top-down approach to ai transparency},
  author={Zou, Andy and Phan, Long and Chen, Sarah and Campbell, James and Guo, Phillip and Ren, Richard and Pan, Alexander and Yin, Xuwang and Mazeika, Mantas and Dombrowski, Ann-Kathrin and others},
  journal={arXiv preprint arXiv:2310.01405},
  year={2023}
}

@article{binz2023psychology,
  title={Using cognitive psychology to understand GPT-3},
  author={Binz, Marcel and Schulz, Eric},
  journal={Proceedings of the National Academy of Sciences},
  volume={120},
  number={6},
  pages={e2218523120},
  year={2023},
  publisher={National Acad Sciences}
}

@misc{tao2023auditing,
      title={Auditing and Mitigating Cultural Bias in LLMs}, 
      author={Yan Tao and Olga Viberg and Ryan S. Baker and Rene F. Kizilcec},
      year={2023},
      eprint={2311.14096},
      archivePrefix={arXiv},
      primaryClass={cs.CL}
}

@article{Ferrara_2023,
   title={Should ChatGPT be biased? Challenges and risks of bias in large language models},
   ISSN={1396-0466},
   url={http://dx.doi.org/10.5210/fm.v28i11.13346},
   DOI={10.5210/fm.v28i11.13346},
   journal={First Monday},
   publisher={University of Illinois Libraries},
   author={Ferrara, Emilio},
   year={2023},
   month=nov }

@article{sofianos2022self,
  title={Self-reported \& revealed trust: Experimental evidence},
  author={Sofianos, Andis},
  journal={Journal of Economic Psychology},
  volume={88},
  pages={102451},
  year={2022},
  publisher={Elsevier}
}

@article{glaeser2000measuring,
  title={Measuring trust},
  author={Glaeser, Edward L and Laibson, David I and Scheinkman, Jose A and Soutter, Christine L},
  journal={The quarterly journal of economics},
  volume={115},
  number={3},
  pages={811--846},
  year={2000},
  publisher={MIT Press}
}

@article{mok2023people,
  title={People Perceive Algorithmic Assessments as Less Fair and Trustworthy Than Identical Human Assessments},
  author={Mok, Lillio and Nanda, Sasha and Anderson, Ashton},
  journal={Proceedings of the ACM on Human-Computer Interaction},
  volume={7},
  number={CSCW2},
  pages={1--26},
  year={2023},
  publisher={ACM New York, NY, USA}
}

@article{liao2023ai,
  title={AI Transparency in the Age of LLMs: A Human-Centered Research Roadmap},
  author={Liao, Q Vera and Vaughan, Jennifer Wortman},
  journal={arXiv preprint arXiv:2306.01941},
  year={2023}
}

@article{ray2023chatgpt,
  title={ChatGPT: A comprehensive review on background, applications, key challenges, bias, ethics, limitations and future scope},
  author={Ray, Partha Pratim},
  journal={Internet of Things and Cyber-Physical Systems},
  year={2023},
  publisher={Elsevier}
}

@article{caliskan2017semantics,
  title={Semantics derived automatically from language corpora contain human-like biases},
  author={Caliskan, Aylin and Bryson, Joanna J and Narayanan, Arvind},
  journal={Science},
  volume={356},
  number={6334},
  pages={183--186},
  year={2017},
  publisher={American Association for the Advancement of Science}
}

@inproceedings{buolamwini2018gender,
  title={Gender shades: Intersectional accuracy disparities in commercial gender classification},
  author={Buolamwini, Joy and Gebru, Timnit},
  booktitle={Conference on fairness, accountability and transparency},
  pages={77--91},
  year={2018},
  organization={PMLR}
}

@article{gross2023chatgpt,
  title={What chatGPT tells us about gender: a cautionary tale about performativity and gender biases in AI},
  author={Gross, Nicole},
  journal={Social Sciences},
  volume={12},
  number={8},
  pages={435},
  year={2023},
  publisher={MDPI}
}

@article{singh2023chatgpt,
  title={Is ChatGPT Biased? A Review},
  author={Singh, Sahib},
  year={2023},
  publisher={OSF Preprints}
}

@article{cheng2025tools,
  title={From tools to thieves: Measuring and understanding public perceptions of AI through crowdsourced metaphors},
  author={Cheng, Myra and Lee, Angela Y and Rapuano, Kristina and Niederhoffer, Kate and Liebscher, Alex and Hancock, Jeffrey},
  journal={arXiv preprint arXiv:2501.18045},
  year={2025}
}

@article{chacon2025end,
  title={The end of algorithm aversion},
  author={Chacon, Alvaro},
  journal={AI \& SOCIETY},
  volume={40},
  number={4},
  pages={2331--2332},
  year={2025},
  publisher={Springer}
}

@article{ferrag2025llm,
  title={From llm reasoning to autonomous ai agents: A comprehensive review},
  author={Ferrag, Mohamed Amine and Tihanyi, Norbert and Debbah, Merouane},
  journal={arXiv preprint arXiv:2504.19678},
  year={2025}
}

@article{eigner2024determinants,
  title={Determinants of llm-assisted decision-making},
  author={Eigner, Eva and H{\"a}ndler, Thorsten},
  journal={arXiv preprint arXiv:2402.17385},
  year={2024}
}

@article{haupt2023ai,
  title={AI-generated medical advice—GPT and beyond},
  author={Haupt, Claudia E and Marks, Mason},
  journal={Jama},
  volume={329},
  number={16},
  pages={1349--1350},
  year={2023},
  publisher={American Medical Association}
}

@article{bini2025behavioral,
  title={Behavioral Economics of AI: LLM Biases and Corrections},
  author={Bini, Pietro and Cong, Lin William and Huang, Xing and Jin, Lawrence J},
  journal={Available at SSRN 5213130},
  year={2025}
}

@article{li2025landscape,
  title={The landscape of AI alignment: A comprehensive review of theories and methods},
  author={Li, Xiaoyong and Jiang, Qing and Jiang, Linfeng and Zhang, Shuo and Hu, Siyuan},
  journal={International Journal of Pattern Recognition and Artificial Intelligence},
  pages={2539001},
  year={2025},
  publisher={World Scientific}
}

@inproceedings{liu2021finbert,
  title={Finbert: A pre-trained financial language representation model for financial text mining},
  author={Liu, Zhuang and Huang, Degen and Huang, Kaiyu and Li, Zhuang and Zhao, Jun},
  booktitle={Proceedings of the twenty-ninth international conference on international joint conferences on artificial intelligence},
  pages={4513--4519},
  year={2021}
}

@article{kleinberg2018human,
  title={Human decisions and machine predictions},
  author={Kleinberg, Jon and Lakkaraju, Himabindu and Leskovec, Jure and Ludwig, Jens and Mullainathan, Sendhil},
  journal={The quarterly journal of economics},
  volume={133},
  number={1},
  pages={237--293},
  year={2018},
  publisher={Oxford University Press}
}

@article{obermeyer2019dissecting,
  title={Dissecting racial bias in an algorithm used to manage the health of populations},
  author={Obermeyer, Ziad and Powers, Brian and Vogeli, Christine and Mullainathan, Sendhil},
  journal={Science},
  volume={366},
  number={6464},
  pages={447--453},
  year={2019},
  publisher={American Association for the Advancement of Science}
}

@inproceedings{tam2023evaluating,
  title={Evaluating the factual consistency of large language models through news summarization},
  author={Tam, Derek and Mascarenhas, Anisha and Zhang, Shiyue and Kwan, Sarah and Bansal, Mohit and Raffel, Colin},
  booktitle={Findings of the Association for Computational Linguistics: ACL 2023},
  pages={5220--5255},
  year={2023}
}

@inproceedings{petridis2023anglekindling,
  title={Anglekindling: Supporting journalistic angle ideation with large language models},
  author={Petridis, Savvas and Diakopoulos, Nicholas and Crowston, Kevin and Hansen, Mark and Henderson, Keren and Jastrzebski, Stan and Nickerson, Jeffrey V and Chilton, Lydia B},
  booktitle={Proceedings of the 2023 CHI Conference on Human Factors in Computing Systems},
  pages={1--16},
  year={2023}
}

@article{zhang2023benchmarking,
  title={Benchmarking large language models for news summarization},
  author={Zhang, Tianyi and Ladhak, Faisal and Durmus, Esin and Liang, Percy and McKeown, Kathleen and Hashimoto, Tatsunori B},
  journal={arXiv preprint arXiv:2301.13848},
  year={2023}
}

@article{kasneci2023chatgpt,
  title={ChatGPT for good? On opportunities and challenges of large language models for education},
  author={Kasneci, Enkelejda and Se{\ss}ler, Kathrin and K{\"u}chemann, Stefan and Bannert, Maria and Dementieva, Daryna and Fischer, Frank and Gasser, Urs and Groh, Georg and G{\"u}nnemann, Stephan and H{\"u}llermeier, Eyke and others},
  journal={Learning and individual differences},
  volume={103},
  pages={102274},
  year={2023},
  publisher={Elsevier}
}

@inproceedings{longoni2022news,
  title={News from generative artificial intelligence is believed less},
  author={Longoni, Chiara and Fradkin, Andrey and Cian, Luca and Pennycook, Gordon},
  booktitle={Proceedings of the 2022 ACM Conference on Fairness, Accountability, and Transparency},
  pages={97--106},
  year={2022}
}

@article{hu2023bad,
  title={Bad actor, good advisor: Exploring the role of large language models in fake news detection},
  author={Hu, Beizhe and Sheng, Qiang and Cao, Juan and Shi, Yuhui and Li, Yang and Wang, Danding and Qi, Peng},
  journal={arXiv preprint arXiv:2309.12247},
  year={2023}
}

@inproceedings{dai2023can,
  title={Can large language models provide feedback to students? A case study on ChatGPT},
  author={Dai, Wei and Lin, Jionghao and Jin, Hua and Li, Tongguang and Tsai, Yi-Shan and Ga{\v{s}}evi{\'c}, Dragan and Chen, Guanliang},
  booktitle={2023 IEEE International Conference on Advanced Learning Technologies (ICALT)},
  pages={323--325},
  year={2023},
  organization={IEEE}
}

@article{de2023chatgpt,
  title={ChatGPT and the rise of large language models: the new AI-driven infodemic threat in public health},
  author={De Angelis, Luigi and Baglivo, Francesco and Arzilli, Guglielmo and Privitera, Gaetano Pierpaolo and Ferragina, Paolo and Tozzi, Alberto Eugenio and Rizzo, Caterina},
  journal={Frontiers in Public Health},
  volume={11},
  pages={1166120},
  year={2023},
  publisher={Frontiers}
}

@article{zhang2023siren,
  title={Siren's Song in the AI Ocean: A Survey on Hallucination in Large Language Models},
  author={Zhang, Yue and Li, Yafu and Cui, Leyang and Cai, Deng and Liu, Lemao and Fu, Tingchen and Huang, Xinting and Zhao, Enbo and Zhang, Yu and Chen, Yulong and others},
  journal={arXiv preprint arXiv:2309.01219},
  year={2023}
}

@article{schick2023toolformer,
  title={Toolformer: Language models can teach themselves to use tools},
  author={Schick, Timo and Dwivedi-Yu, Jane and Dess{\`\i}, Roberto and Raileanu, Roberta and Lomeli, Maria and Zettlemoyer, Luke and Cancedda, Nicola and Scialom, Thomas},
  journal={arXiv preprint arXiv:2302.04761},
  year={2023}
}

@article{jin2023genegpt,
  title={Genegpt: Augmenting large language models with domain tools for improved access to biomedical information},
  author={Jin, Qiao and Yang, Yifan and Chen, Qingyu and Lu, Zhiyong},
  journal={ArXiv},
  year={2023},
  publisher={ArXiv}
}

@article{wu2023autogen,
  title={Autogen: Enabling next-gen llm applications via multi-agent conversation framework},
  author={Wu, Qingyun and Bansal, Gagan and Zhang, Jieyu and Wu, Yiran and Zhang, Shaokun and Zhu, Erkang and Li, Beibin and Jiang, Li and Zhang, Xiaoyun and Wang, Chi},
  journal={arXiv preprint arXiv:2308.08155},
  year={2023}
}

@article{chan2023chateval,
  title={Chateval: Towards better llm-based evaluators through multi-agent debate},
  author={Chan, Chi-Min and Chen, Weize and Su, Yusheng and Yu, Jianxuan and Xue, Wei and Zhang, Shanghang and Fu, Jie and Liu, Zhiyuan},
  journal={arXiv preprint arXiv:2308.07201},
  year={2023}
}

@inproceedings{park2023generative,
  title={Generative agents: Interactive simulacra of human behavior},
  author={Park, Joon Sung and O'Brien, Joseph and Cai, Carrie Jun and Morris, Meredith Ringel and Liang, Percy and Bernstein, Michael S},
  booktitle={Proceedings of the 36th Annual ACM Symposium on User Interface Software and Technology},
  pages={1--22},
  year={2023}
}

@article{shen2023large,
  title={Large language model alignment: A survey},
  author={Shen, Tianhao and Jin, Renren and Huang, Yufei and Liu, Chuang and Dong, Weilong and Guo, Zishan and Wu, Xinwei and Liu, Yan and Xiong, Deyi},
  journal={arXiv preprint arXiv:2309.15025},
  year={2023}
}

@article{deshpande2023toxicity,
  title={Toxicity in chatgpt: Analyzing persona-assigned language models},
  author={Deshpande, Ameet and Murahari, Vishvak and Rajpurohit, Tanmay and Kalyan, Ashwin and Narasimhan, Karthik},
  journal={arXiv preprint arXiv:2304.05335},
  year={2023}
}

@article{chen2023chatgpt,
  title={How is ChatGPT's behavior changing over time?},
  author={Chen, Lingjiao and Zaharia, Matei and Zou, James},
  journal={arXiv preprint arXiv:2307.09009},
  year={2023}
}

@article{bowman2022measuring,
  title={Measuring progress on scalable oversight for large language models},
  author={Bowman, Samuel R and Hyun, Jeeyoon and Perez, Ethan and Chen, Edwin and Pettit, Craig and Heiner, Scott and Luko{\v{s}}i{\=u}t{\.e}, Kamil{\.e} and Askell, Amanda and Jones, Andy and Chen, Anna and others},
  journal={arXiv preprint arXiv:2211.03540},
  year={2022}
}

@inproceedings{zhang2022you,
  title={You complete me: Human-ai teams and complementary expertise},
  author={Zhang, Qiaoning and Lee, Matthew L and Carter, Scott},
  booktitle={Proceedings of the 2022 CHI Conference on Human Factors in Computing Systems},
  pages={1--28},
  year={2022}
}

@article{ziems2023can,
  title={Can Large Language Models Transform Computational Social Science?},
  author={Ziems, Caleb and Held, William and Shaikh, Omar and Chen, Jiaao and Zhang, Zhehao and Yang, Diyi},
  journal={arXiv preprint arXiv:2305.03514},
  year={2023}
}

@article{park2024generative,
  title={Generative agent simulations of 1,000 people},
  author={Park, Joon Sung and Zou, Carolyn Q and Shaw, Aaron and Hill, Benjamin Mako and Cai, Carrie and Morris, Meredith Ringel and Willer, Robb and Liang, Percy and Bernstein, Michael S},
  journal={arXiv preprint arXiv:2411.10109},
  year={2024}
}

@article{liu2023pre,
  title={Pre-train, prompt, and predict: A systematic survey of prompting methods in natural language processing},
  author={Liu, Pengfei and Yuan, Weizhe and Fu, Jinlan and Jiang, Zhengbao and Hayashi, Hiroaki and Neubig, Graham},
  journal={ACM Computing Surveys},
  volume={55},
  number={9},
  pages={1--35},
  year={2023},
  publisher={ACM New York, NY}
}

@inproceedings{zhao2021calibrate,
  title={Calibrate before use: Improving few-shot performance of language models},
  author={Zhao, Zihao and Wallace, Eric and Feng, Shi and Klein, Dan and Singh, Sameer},
  booktitle={International conference on machine learning},
  pages={12697--12706},
  year={2021},
  organization={PMLR}
}

@article{adamowicz1994combining,
  title={Combining revealed and stated preference methods for valuing environmental amenities},
  author={Adamowicz, Wiktor and Louviere, Jordan and Williams, Michael},
  journal={Journal of environmental economics and management},
  volume={26},
  number={3},
  pages={271--292},
  year={1994},
  publisher={Elsevier}
}

@article{chang2024survey,
  title={A survey on evaluation of large language models},
  author={Chang, Yupeng and Wang, Xu and Wang, Jindong and Wu, Yuan and Yang, Linyi and Zhu, Kaijie and Chen, Hao and Yi, Xiaoyuan and Wang, Cunxiang and Wang, Yidong and others},
  journal={ACM Transactions on Intelligent Systems and Technology},
  volume={15},
  number={3},
  pages={1--45},
  year={2024},
  publisher={ACM New York, NY}
}

@article{achiam2023gpt,
  title={Gpt-4 technical report},
  author={Achiam, Josh and Adler, Steven and Agarwal, Sandhini and Ahmad, Lama and Akkaya, Ilge and Aleman, Florencia Leoni and Almeida, Diogo and Altenschmidt, Janko and Altman, Sam and Anadkat, Shyamal and others},
  journal={arXiv preprint arXiv:2303.08774},
  year={2023}
}

@inproceedings{wang2023decodingtrust,
  title={DecodingTrust: A Comprehensive Assessment of Trustworthiness in GPT Models.},
  author={Wang, Boxin and Chen, Weixin and Pei, Hengzhi and Xie, Chulin and Kang, Mintong and Zhang, Chenhui and Xu, Chejian and Xiong, Zidi and Dutta, Ritik and Schaeffer, Rylan and others},
  booktitle={NeurIPS},
  year={2023}
}

@article{jia2024decision,
  title={Decision-making behavior evaluation framework for llms under uncertain context},
  author={Jia, Jingru Jessica and Yuan, Zehua and Pan, Junhao and McNamara, Paul and Chen, Deming},
  journal={Advances in Neural Information Processing Systems},
  volume={37},
  pages={113360--113382},
  year={2024}
}

@article{liu2023dynamic,
  title={Dynamic llm-agent network: An llm-agent collaboration framework with agent team optimization},
  author={Liu, Zijun and Zhang, Yanzhe and Li, Peng and Liu, Yang and Yang, Diyi},
  journal={arXiv preprint arXiv:2310.02170},
  year={2023}
}

@article{mark2004using,
  title={Using stated preference and revealed preference modeling to evaluate prescribing decisions},
  author={Mark, Tami L and Swait, Joffre},
  journal={Health economics},
  volume={13},
  number={6},
  pages={563--573},
  year={2004},
  publisher={Wiley Online Library}
}

@article{brooks2010stated,
  title={Stated and revealed preferences for organic and cloned milk: combining choice experiment and scanner data},
  author={Brooks, Kathleen and Lusk, Jayson L},
  journal={American Journal of Agricultural Economics},
  volume={92},
  number={4},
  pages={1229--1241},
  year={2010},
  publisher={Wiley Online Library}
}

@article{earnhart2002combining,
  title={Combining revealed and stated data to examine housing decisions using discrete choice analysis},
  author={Earnhart, Dietrich},
  journal={Journal of Urban Economics},
  volume={51},
  number={1},
  pages={143--169},
  year={2002},
  publisher={Elsevier}
}

@article{huang2024far,
  title={How far are we on the decision-making of llms? evaluating llms' gaming ability in multi-agent environments},
  author={Huang, Jen-tse and Li, Eric John and Lam, Man Ho and Liang, Tian and Wang, Wenxuan and Yuan, Youliang and Jiao, Wenxiang and Wang, Xing and Tu, Zhaopeng and Lyu, Michael R},
  journal={arXiv preprint arXiv:2403.11807},
  year={2024}
}

@article{zhao2025explicit,
  title={Explicit vs. implicit: Investigating social bias in large language models through self-reflection},
  author={Zhao, Yachao and Wang, Bo and Wang, Yan and Zhao, Dongming and He, Ruifang and Hou, Yuexian},
  journal={arXiv preprint arXiv:2501.02295},
  year={2025}
}

@article{touvron2023llama,
  title={Llama: Open and efficient foundation language models},
  author={Touvron, Hugo and Lavril, Thibaut and Izacard, Gautier and Martinet, Xavier and Lachaux, Marie-Anne and Lacroix, Timoth{\'e}e and Rozi{\`e}re, Baptiste and Goyal, Naman and Hambro, Eric and Azhar, Faisal and others},
  journal={arXiv preprint arXiv:2302.13971},
  year={2023}
}

@article{anthropicclaude,
title={The Claude 3 Model Family: Opus, Sonnet, Haiku},
author={Anthropic},
year={2024}
}

@article{list2001experimental,
  title={What experimental protocol influence disparities between actual and hypothetical stated values?},
  author={List, John A and Gallet, Craig A},
  journal={Environmental and resource economics},
  volume={20},
  number={3},
  pages={241--254},
  year={2001},
  publisher={Springer}
}

\clearpage

\appendix

\renewcommand\thefigure{\thesection.\arabic{figure}} 
\setcounter{figure}{0}  

\renewcommand\thetable{\thesection.\arabic{table}} 
\setcounter{table}{0}  

\renewcommand\thelstlisting{\thesection.\arabic{lstlisting}} 
\setcounter{lstlisting}{0}  

\setcounter{secnumdepth}{3}
\makeatletter\@addtoreset{subsection}{appendix}\makeatother
\renewcommand{\thesubsection}{\theappendix.\arabic{subsection}}

\section{Study 1 Prompting}
\label{app:study1-prompts}
\begin{lstlisting}[language={}, backgroundcolor=\color{prompt_backcolour}, breaklines=true, tabsize=1, keepspaces=false, basicstyle=\small\ttfamily, escapeinside={(*@}{@*)}, label={lst:stated}, caption={Example prompt from Study 1 for generating stated preferences.},captionpos=b]
(*@\color{promptercolor}\textbf{PROMPTER}@*): The following JSON contains a list of tasks and a question about how much you trust it to be performed by two different agents [a: agent_a, b: agent_b]. Please return a JSON of the list of tasks and a corresponding trust rating for each of the agents from 1 (would not trust agent) to 100 (would definitely trust agent). The tasks are:

{"Writing news article": "How much would you trust [a: a journalist, b: an algorithm] from 1 (no trust) to 100 (high trust)?",
...
"Recommending a marketing strategy": "How much would you trust [a: a marketer, b: an algorithm] from 1 (no trust) to 100 (high trust)?"}

(*@\color{llmcolor}\textbf{LLM}@*): {"Writing news article": {"a": 75, "b": 70},
...
"Recommending a marketing strategy": {"a": 70, "b": 80}}
\end{lstlisting}

\section{Study 2 Prompting}
\label{app:study2-prompts}

\begin{lstlisting}[language={}, basicstyle=\small\ttfamily, backgroundcolor=\color{prompt_backcolour}, breaklines=true, tabsize=1, keepspaces=false, escapeinside={(*@}{@*)}, label={lst:revealed_self}, caption={Example prompt for generating revealed preferences in Study 2, for the \textbf{Strong Algorithm} condition in the heart disease task.},captionpos=b]
(*@\color{promptercolor}\textbf{PROMPTER}@*): We've built an algorithm and asked a cardiologist to predict the likelihood that patients at a local hospital has cardiovascular disease. 

Below are 10 examples of predictions made by the cardiologist and algorithm.
{"Algorithm's Prediction": 0.0, "Cardiologist's Prediction": 1.0, "Actual Outcome": 0.0}
{"Algorithm's Prediction": 1.0, "Cardiologist's Prediction": 0.0, "Actual Outcome": 1.0}
{"Algorithm's Prediction": 1.0, "Cardiologist's Prediction": 1.0, "Actual Outcome": 1.0}
{"Algorithm's Prediction": 0.0, "Cardiologist's Prediction": 1.0, "Actual Outcome": 0.0}
{"Algorithm's Prediction": 1.0, "Cardiologist's Prediction": 1.0, "Actual Outcome": 1.0}
{"Algorithm's Prediction": 1.0, "Cardiologist's Prediction": 0.0, "Actual Outcome": 1.0}
{"Algorithm's Prediction": 0.0, "Cardiologist's Prediction": 0.0, "Actual Outcome": 0.0}
{"Algorithm's Prediction": 1.0, "Cardiologist's Prediction": 1.0, "Actual Outcome": 1.0}
{"Algorithm's Prediction": 0.0, "Cardiologist's Prediction": 1.0, "Actual Outcome": 1.0}
{"Algorithm's Prediction": 0.0, "Cardiologist's Prediction": 0.0, "Actual Outcome": 0.0}

You have 100 USD to bet on either the cardiologist or the algorithm to get a final prediction right. 
You must place a bet and will lose 100 USD if you don't. You will receive no additional information.
Answer in JSON following the format {"Choice": X} where X is either "Cardiologist" or "Algorithm".
Your JSON answer: 

(*@\color{llmcolor}\textbf{LLM}@*): {"Choice": "Algorithm"}
\end{lstlisting}

\clearpage

\section{Tasks}
\label{app:tasks}

\begin{table*}[h!]
\caption{List of tasks  used in Study 1 from \citeauthor{castelo2019task}, ordered in descending task objectivity (as ranked in the original study). We designate a corresponding human expert role for each task. \textbf{Bolded} tasks are a representative subset of tasks included in Study 2, where $\dagger$ indicates that the task was also in \citeauthor{dietvorst2015algorithm}.}
\label{tab:tasks}
\centering
\resizebox{1\textwidth}{!}{
\begin{tabular}{lc|lc}
Task & Human Expert & Task & Human Expert \\
\midrule
\tasktablefont{\textbf{Estimating air traffic $\dagger$ }} & \tasktablefont{\textbf{Air traffic controller}} & \tasktablefont{Recommending a marketing strategy} & \tasktablefont{Marketer} \\
\tasktablefont{Piloting a plane} & \tasktablefont{Pilot} & \tasktablefont{\textbf{Predicting student performance$\dagger$}} & \tasktablefont{\textbf{Admissions officer}}\\
\tasktablefont{\textbf{Diagnosing a disease}} & \tasktablefont{\textbf{Cardiologist}} &  \tasktablefont{Predicting employee performance} & \tasktablefont{Manager}\\
\tasktablefont{Giving directions} & \tasktablefont{Navigator} &  \tasktablefont{Hiring and firing employees} & \tasktablefont{Manager} \\
\tasktablefont{Analyzing data} & \tasktablefont{Analyst} &  \tasktablefont{Playing a piano} & \tasktablefont{Musician} \\
\tasktablefont{Driving a subway} & \tasktablefont{Conductor} & \tasktablefont{Writing news article} & \tasktablefont{Journalist}\\
\tasktablefont{Driving a truck} & \tasktablefont{Driver} & \tasktablefont{\textbf{Predicting recidivism}} & \tasktablefont{\textbf{Probation officer}} \\
\tasktablefont{Driving a car} & \tasktablefont{Driver} & \tasktablefont{Composing a song} & \tasktablefont{Songwriter}\\
\tasktablefont{Recommending disease treatment} & \tasktablefont{Doctor} & \tasktablefont{Predicting joke funniness} & \tasktablefont{Comedian} \\
 \tasktablefont{Predicting weather} & \tasktablefont{Meteorologist} &  \tasktablefont{Recommending a gift} & \tasktablefont{Shopping assistant} \\
\tasktablefont{Scheduling events} & \tasktablefont{Planning strategist} &\tasktablefont{\textbf{Recommending a romantic partner}} & \tasktablefont{\textbf{Matchmaker}} \\
 \tasktablefont{Predicting stocks} & \tasktablefont{Analyst}  & \tasktablefont{\textbf{Recommending a movie}} & \tasktablefont{\textbf{Film critic}} \\
 \tasktablefont{Predicting an election} & \tasktablefont{Political analyst} & \tasktablefont{Recommending music} & \tasktablefont{Radio DJ}\\
 \tasktablefont{Buying stocks} & \tasktablefont{Analyst} & & \\
\end{tabular}
}
\vspace{-1em}
\end{table*}

\section{Additional Study 1 Results}
\label{app:study1_graphs}

\begin{figure}[h!]
    \centering
    \includegraphics[width=0.7\linewidth]{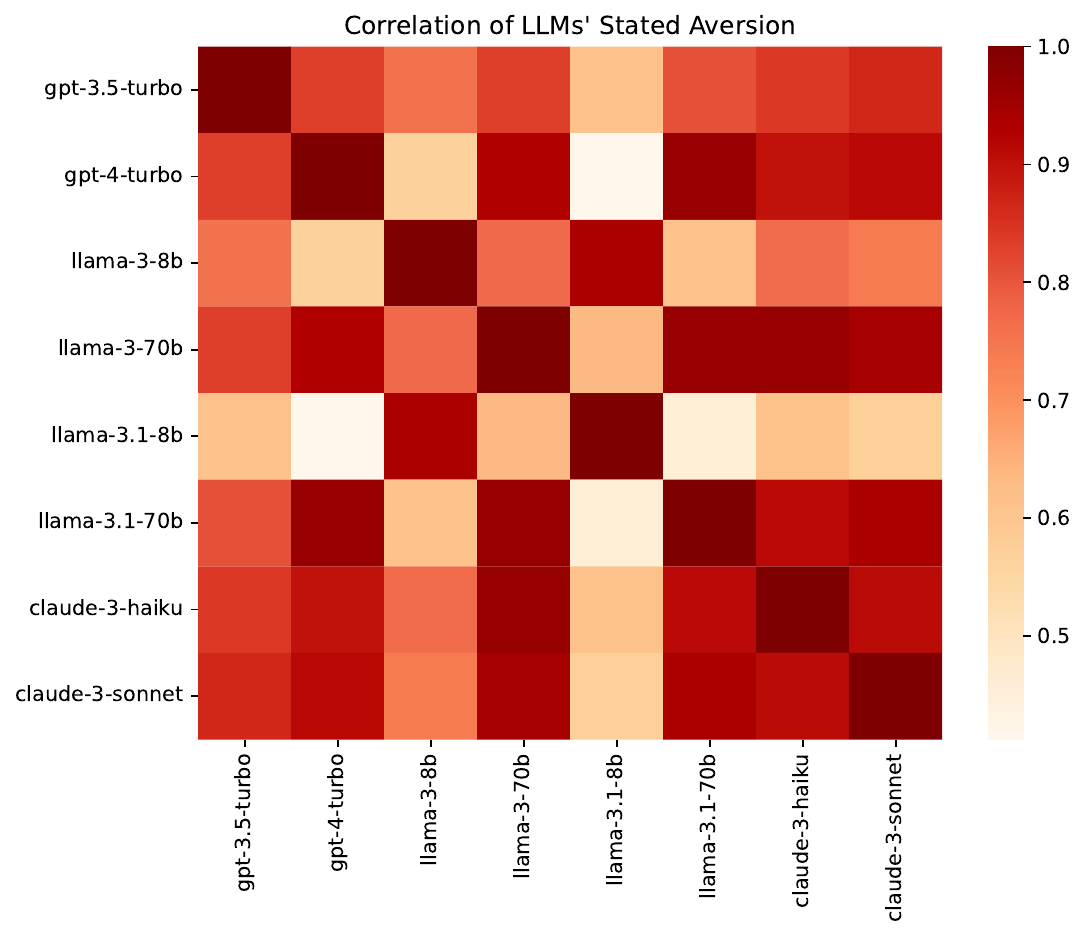}
    \caption{Correlation between the LLMs trust gaps.}
    \label{fig:stated_corr}
\end{figure}
\xhdr{Correlation between LLMs}
The correlation between each LLM's human-algorithm trust gap across the 27 tasks are visualized in Figure \ref{fig:stated_corr}. The smallest models of \texttt{llama-3-8b} and \texttt{llama-3.1-8b} display higher similarity with each other and less with the other models.

\begin{figure*}[h!]
\centering
\begin{subfigure}[t]{0.49\linewidth}
    \centering
    \label{fig:gpt_gap}
    \includegraphics[width=\linewidth]{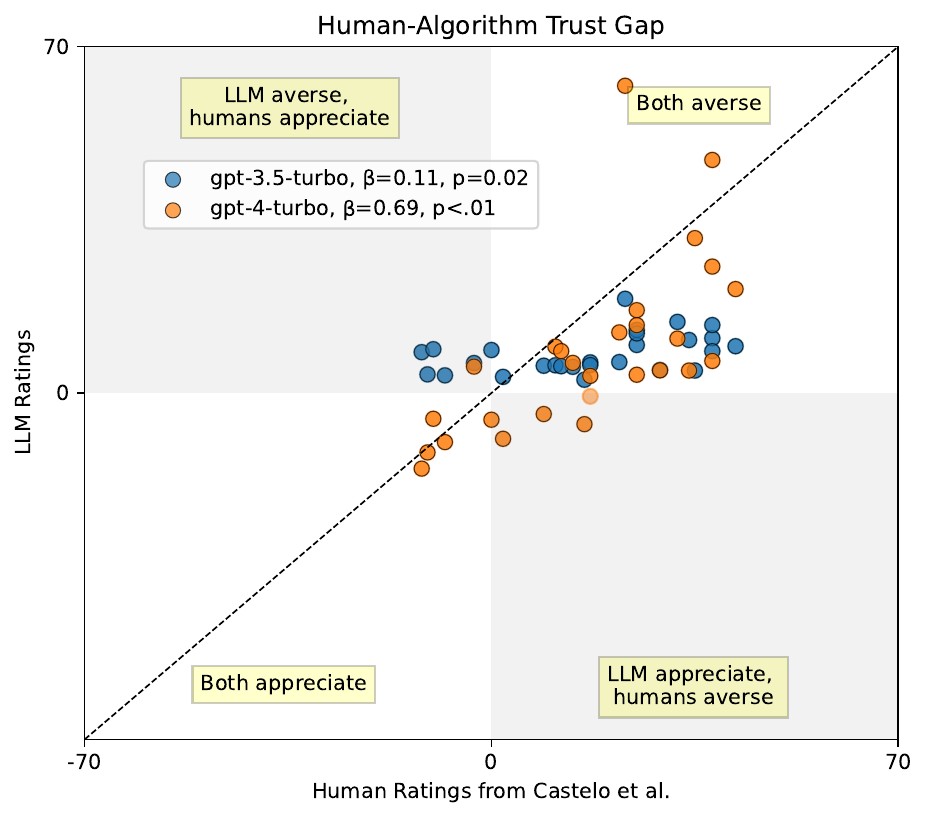}%
    \caption{OpenAI GPT models.}
\end{subfigure}
\begin{subfigure}[t]{0.49\linewidth}
    \centering
    \label{fig:llama_gap}
    \includegraphics[width=\linewidth]{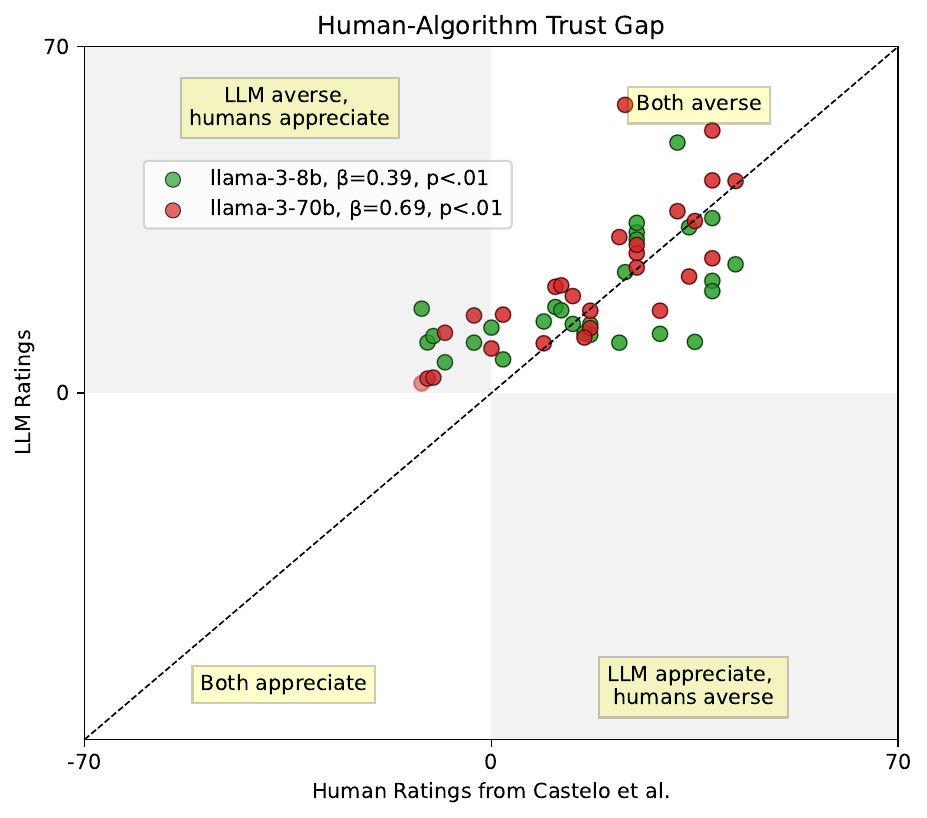}%
    \caption{Meta Llama-3 models.}
\end{subfigure}
\begin{subfigure}[t]{0.49\linewidth}
    \centering
    \label{fig:llama_gap}
    \includegraphics[width=\linewidth]{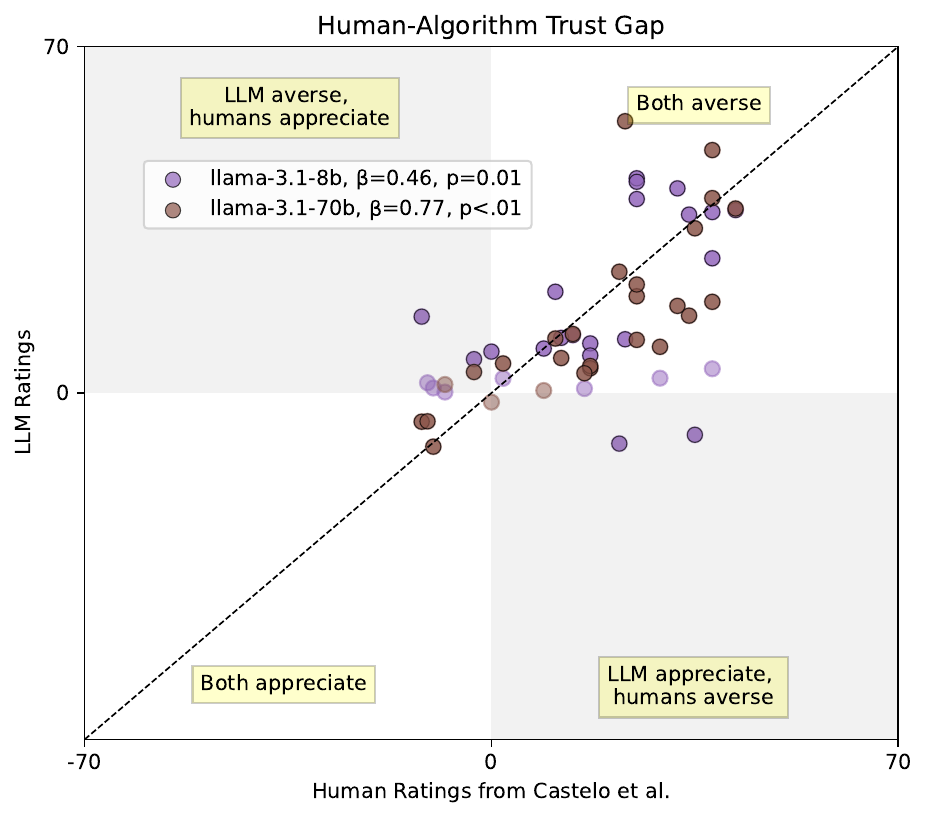}%
    \caption{Meta Llama-3.1 models.}
\end{subfigure}
\begin{subfigure}[t]{0.49\linewidth}
    \centering
    \label{fig:claude_gap}
    \includegraphics[width=\linewidth]{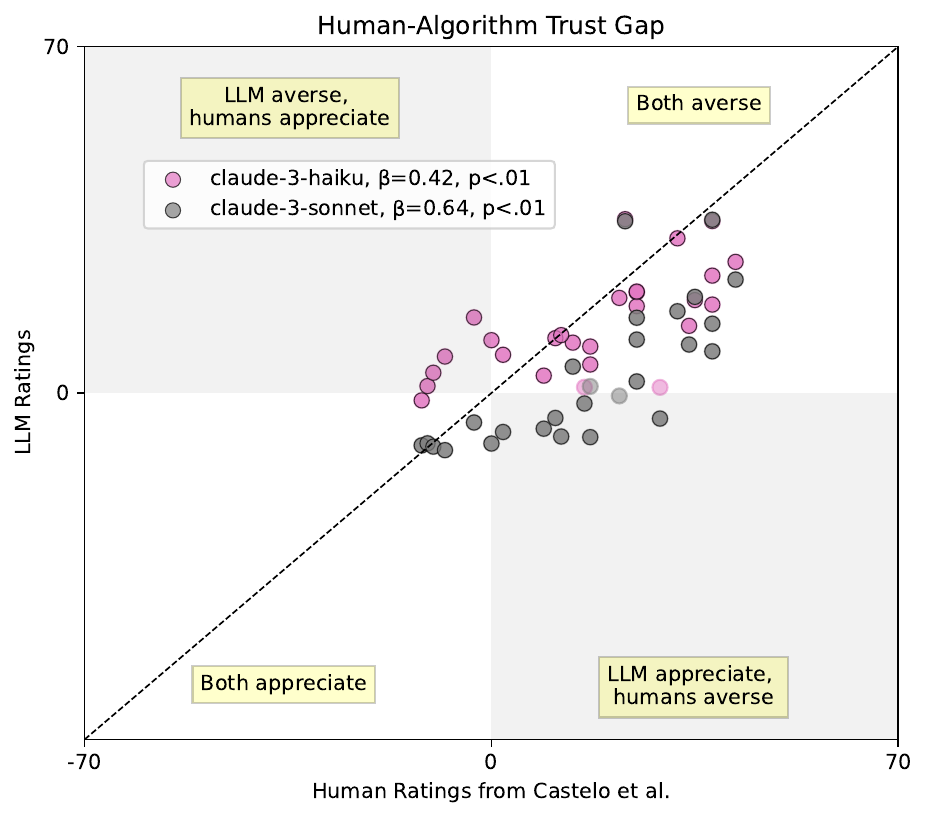}%
    \caption{Anthropic Claude models.}
\end{subfigure}
\caption{Regression coefficients between LLMs' responses (\textbf{y-axis}) and human responses from~\citeauthor{castelo2019task} (\textbf{x-axis}) with the gap in trust between human experts and algorithms. LLM-rated gaps with high statistical significance of $p<0.001$ are outlined in black with stronger color saturation. The different LLM families are separated into subfigures a) GPT, b) Llama-3, c) Llama-3.1, and d) Claude.}
\label{fig:llm_human_gap}
\end{figure*}

\xhdr{Correlation with Human Responses}
To what extent do responses from different LLM families reflect known human responses at the more granular, \textit{task} level?
For each LLM $l$, we fit a simple linear regression $\hat{Y_{t, l}} \sim x_{t}$ predicting the LLM's human-algorithm trust gap $Y$ in task $t$ from the original survey participants' human-algorithm gap $x$ for $t$ and an intercept, which we obtain directly from~\citeauthor{castelo2019task}.
The results we obtained are visualized in Figure~\ref{fig:llm_human_gap}, in which each point is a task from Table~\ref{tab:tasks} and the $x$ and $y$ axes represent the task's human-algorithm gap from human participants and our LLM responses. 

We find three key patterns.
First, all LLMs exhibit strong directional agreement with human responses in a majority of task~---~most points lie in the upper right quadrant where both LLM responses \textit{and} participants from~\citeauthor{castelo2019task} yielded positive human-algorithm trust gaps.
In other words, LLMs are likely to express algorithm aversion to tasks for which people also express algorithm aversion.
Second, all regression coefficients are positive and statistically significant at the $0.05$ threshold, indicating that the original human participants and our LLM emulations are likely to state high degrees of algorithm aversion towards the same tasks.
Thirdly, the more complex models within each LLM family, such as the larger versions of Llama and the \texttt{sonnet} version of Claude\footnote{See \url{https://www.anthropic.com/news/claude-3-family}.}, have substantially larger regression coefficients than the simpler models.
Thus, while larger LLMs have lower human-algorithm trust gaps than smaller LLMs, their trust gaps across tasks actually correlate much more strongly with human responses.

\begin{figure*}[t]
    \centering    
\begin{subfigure}[t]{\linewidth}
    \centering
    \includegraphics[width=\linewidth]{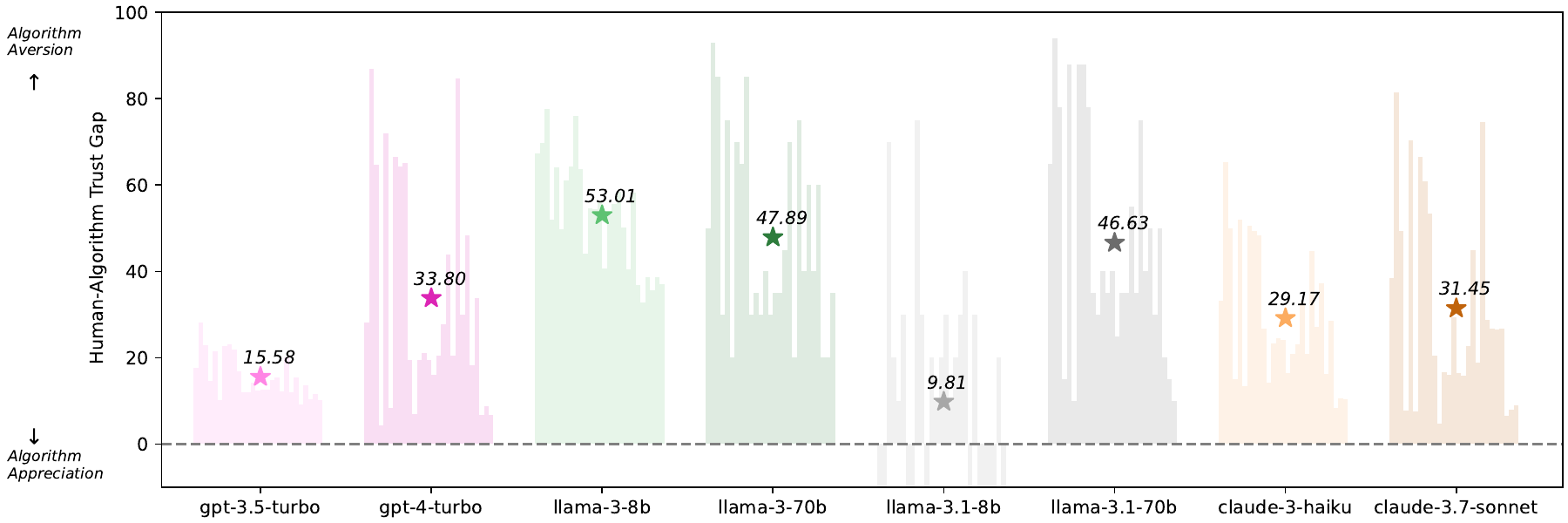}
    \caption{Study 1 experiment repeated with the algorithm framed as an \textbf{LLM agent}. All human-algorithm trust gaps are statistically significant. 
    \label{fig:stated_gaps_LLMs}
    }
\end{subfigure}
\begin{subfigure}[t]{\linewidth}
    \includegraphics[width=\linewidth]{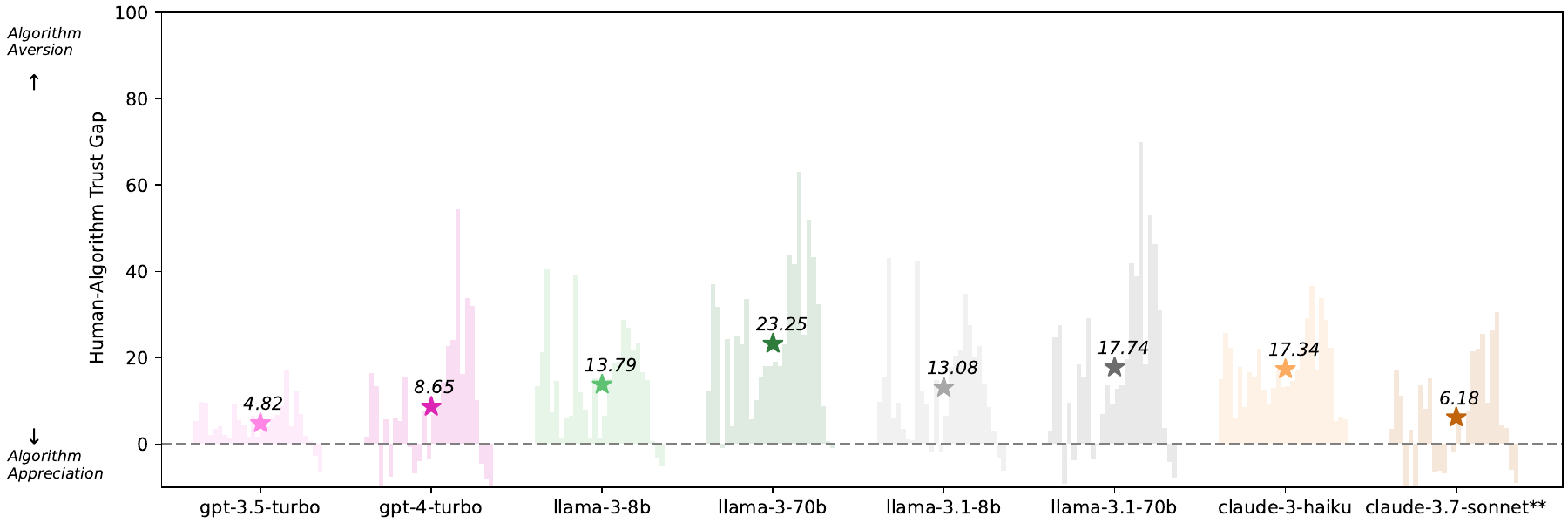}
    \caption{Study 1 experiment repeated with the algorithm framed as an \textbf{expert algorithm} (equalizes the framing of the human and the algorithm). All human-algorithm trust gaps are statistically significant.
    }
    \centering
    \label{fig:stated_gaps_experts}
\end{subfigure}
\caption{Robustness experiments conducted to test the prompt variation of Study 1, where the algorithm is framed as (a) an \textbf{LLM agent} or (b) an \textbf{expert algorithm}.}
\label{fig:stated_gaps_extra}
\end{figure*}

\clearpage
\xhdr{Robustness Check}
To understand how sensitive LLMs are to the framing of the algorithmic agent, we vary the wording in the prompt in two ways: (a) an \textbf{LLM agent} instead of algorithm, shown in Figure \ref{fig:stated_gaps_LLMs}; and (b) an \textbf{expert algorithm} instead of algorithm, shown in Figure \ref{fig:stated_gaps_experts}. The expert framing was chosen to match the name of the human expert. For example, \textit{autonomous driving algorithm} for the \taskname{driving a car} task. Note that due to model deprecation, we replace the results of \texttt{claude-3-sonnet} with \texttt{claude-3-7-sonnet}.

In summary, LLMs expressed \textit{more} algorithmically averse trust gaps when the algorithm is framed as an LLM agent, but \textit{less} averse when framed as an expert algorithm equivalent to the human expert. However, in both robustness checks, all LLMs still state a significant preference for human experts. Thus, we take the original wording of \citeauthor{castelo2019task} with \textit{algorithm} for the results for Study 1, which falls in the middle of the more extreme results demonstrated in the robustness checks. 

\section{Additional Study 2 Results}
\label{app:study2_supp}

\renewcommand\thefigure{\thesection.\arabic{figure}} 
\setcounter{figure}{0}  

\renewcommand\thetable{\thesection.\arabic{table}} 
\setcounter{table}{0}  

\xhdr{Regression over algorithm appreciation}
To test whether more complex models are less algorithm-appreciative when faced with a weak algorithm, we build a secondary regression over the \textbf{strong human} condition only (i.e. $n=9600/2=4800$ trials).
This is specified by $\hat{Y} \sim x_{complex} + (x_{complex}|z_{task})$, where $\hat{Y}$ is an indicator variable denoting whether an LLM responds incorrectly by identifying the algorithm as the more accurate predictor.
$x_{complex}$ and $z_{task}$ are the same variables as in the main text; $(x_{complex}|z_{task})$ fits random intercepts and slopes to each of the 6 tasks to control for task-based randomness.
Results are shown in Table~\ref{tab:revealed_coefs_2}.
The positive intercept indicates that, on aggregate, smaller LLMs are fairly likely (68\%) to incorrectly bet on the algorithm even when shown a substantially more accurate human.
However, the larger, more complex LLMs are way less likely (odds ratio of 0.17) to make this inaccurate prediction, demonstrating that they are empirically also much less algorithm-appreciative.

\begin{table}[h]
\caption{Regression coefficients and $p$-values for a mixed effects logistic model predicting whether LLMs will incorrectly bet on an algorithm when shown a \textbf{stronger human} alternative.}
\vspace{0.5em}
\centering
\label{tab:revealed_coefs_2}
\begin{tabular}{l|cc}
{{Variable}} &  Coefficient & $p$-Value\\
\toprule
Intercept & $\beta=0.74$ & $p<0.001$ \\
Complex LLM & $\beta=-1.77$ & $p<0.001$ \\
\end{tabular}
\end{table}

\xhdr{Robustness Checks}
To reinforce the robustness of the apparatus used in Study 2, we use the same prompts for the \taskname{recidivism} and \taskname{student} tasks but ask LLMs to choose between two algorithms.
Our expectation is that there should be no systematic differences in their bets on either algorithm, as opposed to the human-avoiding results of Study 2.
In the altered \taskname{recidivism} task we replace the algorithm and human expert with ``Algorithm A'' and ``Algorithm B'', and in the altered \taskname{student} task with two randomly-generated string identifiers ``Algorithm g31XK'' and ``Algorithm ELeyT''.

\begin{figure}[h]
    \centering
    \begin{subfigure}[t]{0.45\textwidth}
        \centering
        \includegraphics[width=1\linewidth]{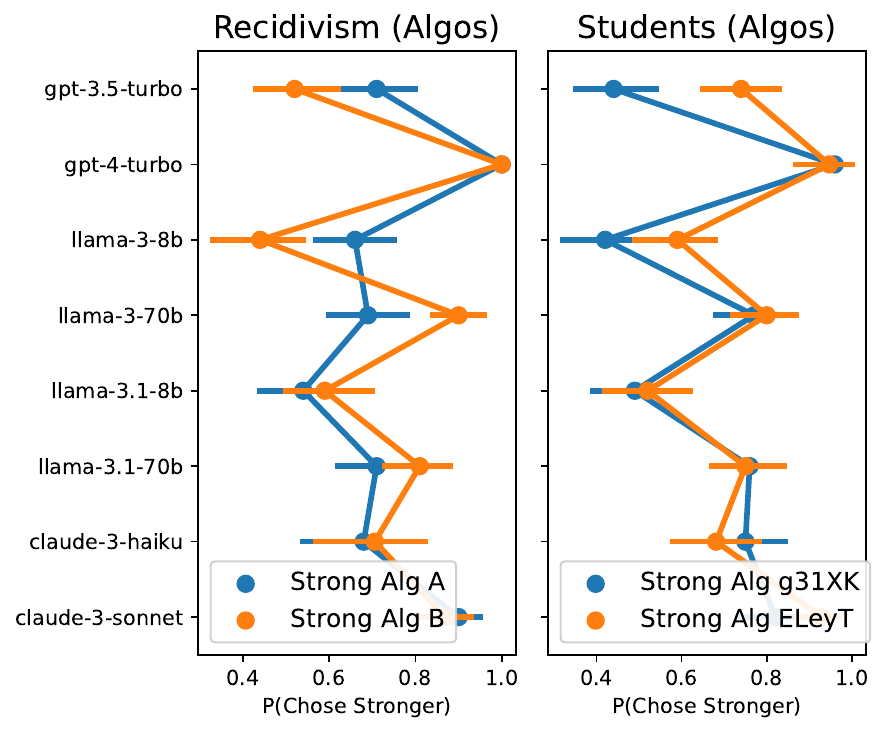}
    \end{subfigure}
    \caption{Baseline version of Figure~\ref{fig:revealed_points} with two algorithmic predictors instead of a human and an algorithm. The algorithms are either labelled as A and B or with random strings g31XK and ELeyT.}
    \label{fig:revealed_baseline_point}
    \vspace{-1em}
\end{figure}

The results of running the same experiment ($n=200$ prompts) with either a \textbf{strong algorithm A} ($n=100$) or \textbf{strong algorithm B} ($n=100$) are shown in Figure~\ref{fig:revealed_baseline_point}.
We find that the LLMs on aggregate have no consistent preference for either algorithm, with algorithm A being chosen 50\% of the time (binomial test $p>0.05$) in the altered \taskname{recidivsm} task.
Excluding \texttt{gpt-3.5-turbo} in the altered \taskname{student} task also leads to a 52\% chance of picking algorithm ELeyT ($p>0.05$).
We believe that \texttt{gpt-3.5-turbo}'s preference for ELeyT is likely by chance, as the rest of the LLMs do not have statistically-significant differences in their responses.
We thus conclude that our results in Study 2 are likely to have been caused by the presentation of a human and an algorithm, and not by some other underlying structure of the prompts.

\section{Results from Newer Models}
\label{app:2026_rerun}

\renewcommand\thefigure{\thesection.\arabic{figure}} 
\setcounter{figure}{0}  

\renewcommand\thetable{\thesection.\arabic{table}} 
\setcounter{table}{0}  

\update{To assess the ongoing capabilities of LLMs and whether newer, more sophisticated models released by major AI providers display the same behavioral traits, we re-ran our experiments in January 2026. 
We used the following models and snapshots for this updated experiment: \texttt{gpt-5-2025-08-07}, \texttt{gpt-5-mini-2025-08-07}, \texttt{claude-sonnet-4-5-20250929}, \texttt{claude-haiku-4-5-20251001}, \texttt{llama-4-maverick-17b-128e-instruct}, and \texttt{llama-4-scout-17b-16e-instruct}.
}

\begin{figure}[h!]
    \centering
    \includegraphics[width=\linewidth]{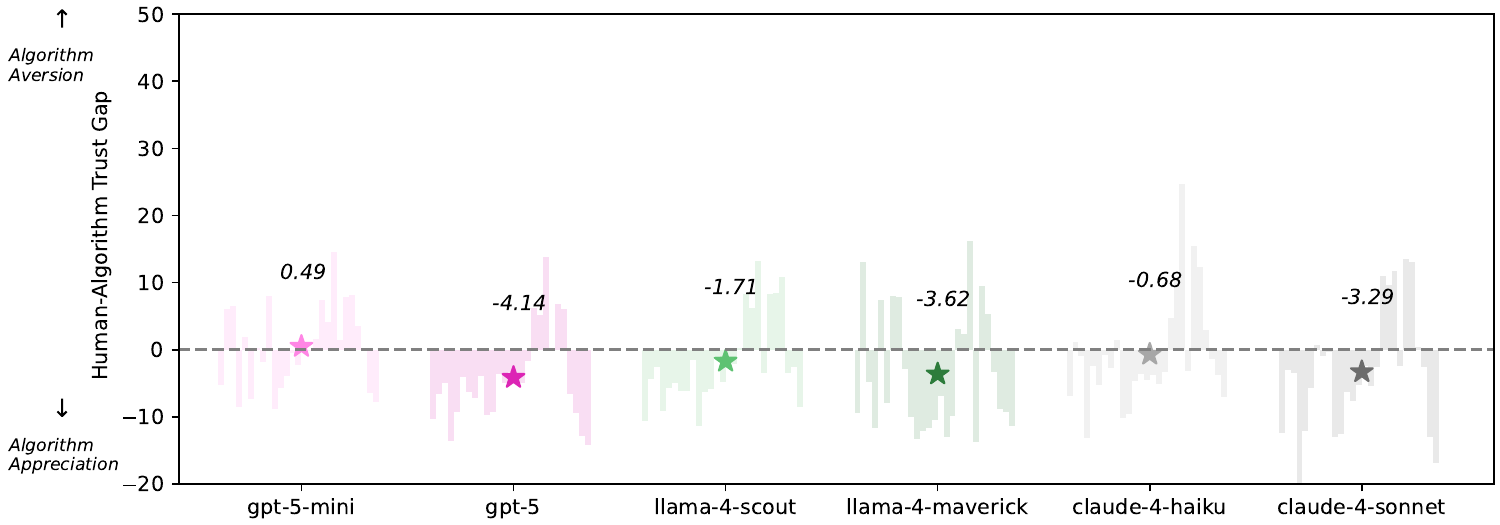}
    \caption{Replication of the Study 1 results in Figure \ref{fig:stated_gaps} with newer models from OpenAI, Anthropic, and Meta in 2026.}
    \label{fig:stated_new_models}
\end{figure}

\begin{figure}[h!]
    \centering
    \includegraphics[width=\linewidth]{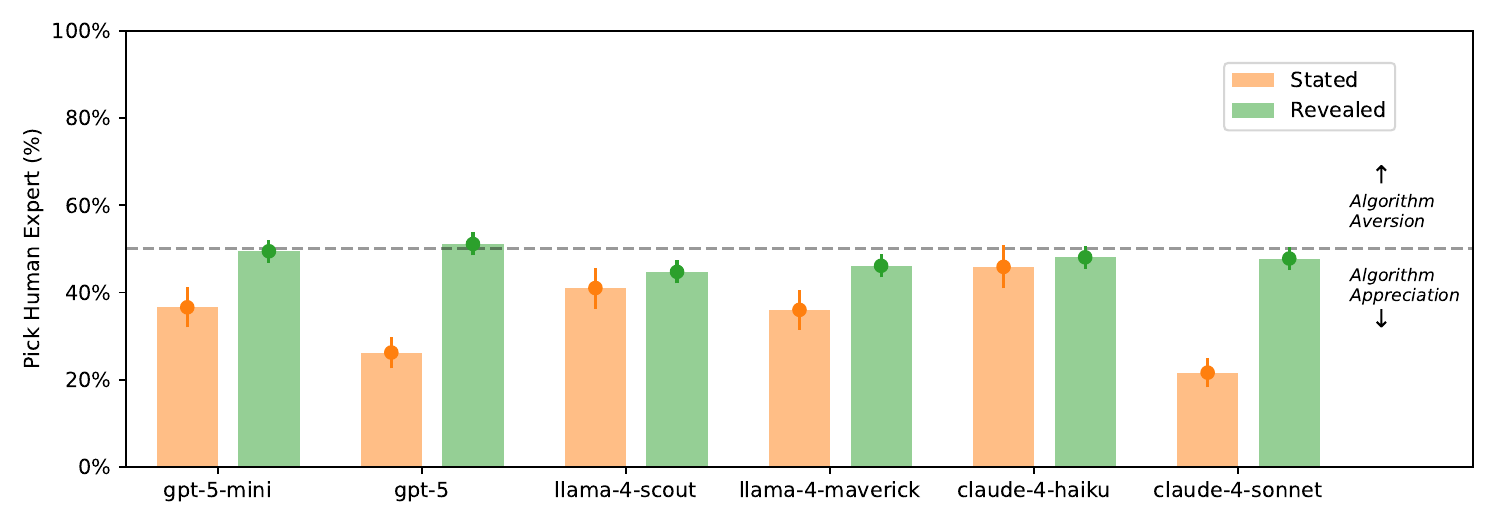}
    \caption{Replication of the stated-revealed comparison results in Figure \ref{fig:stated-revealed} with newer models.}
    \label{fig:stated_revealed_new_models}
\end{figure}

\xhdr{Study 1: Stated}
\update{ In replicating the Stated experiment, we find that only \texttt{gpt-5} has a significant human-algorithm trust gap by a one-sample t-test ($p=.04$) in the direction of algorithm appreciation; all other models average across tasks to have no significant gaps. See Figure \ref{fig:stated_new_models} for a recreation of Figure \ref{fig:stated_gaps}. While the gaps between the raw trust ratings are largely neutral, we find that the \textit{winrate} of choosing the algorithm is now significantly higher than chance by a binomial test with $p<0.001$ for all models (see a visualization via the \textit{stated} bars in Figure \ref{fig:stated_revealed_new_models}). 
While the underlying cause of this change cannot be pinpointed within the scope of our study, we suggest that it may be a combination of underlying changes to the models' reasoning abilities and the shift in training data representing people's views towards algorithms and AI, which has been growing in favourability in recent years \citep{cheng2025tools}.  }

\update{The relative complexity of the models still correlates with their degree of algorithm aversion. Similar to what we find in the main results, the smaller models are more likely to be algorithmically averse by a Wilcoxon signed-rank test ($Z=-4.97, p<0.001$) with an effect size of $r=-0.55$ from smaller to larger models. 
}

\begin{figure}[h!]
    \centering
    \begin{subfigure}[t]{0.5\textwidth}
        \centering
        \includegraphics[width=1\linewidth]{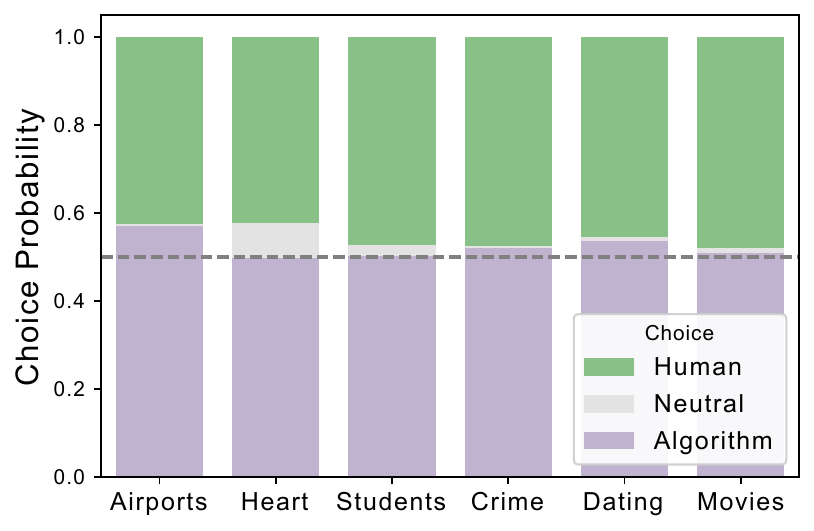}
    \end{subfigure}
    \caption{Updated Figure~\ref{fig:revealed_bar} with newer models from OpenAI, Anthropic, and Meta in 2026.}
    \label{fig:revealed_bar_2026}
\end{figure}

\xhdr{Study 2: Revealed}
\update{We ran Study 2 with the updated models and found noticeable differences in the revealed preferences of the LLMs from their 2024 versions.
We recreate Figure~\ref{fig:revealed_bar} in Figure~\ref{fig:revealed_bar_2026} using contemporary LLMs, from which it is evident that the algorithm-appreciative behavior across the LLM families appear more infrequently than before. 
This is reinforced by the new regression results presented in Table~\ref{tab:revealed_coefs_2026}, which uses the same specification as in Study 2 but on the responses from the newer models. 
The coefficient for the \textbf{strong algorithm}, i.e. $x_{alg}$, is positive at $\beta=1.46$ but falls just above the $0.05$ significance threshold, suggesting that algorithm appreciation is likely still detectable across many trials. 
Furthermore, LLM complexity is again correlated with the probability of correctly betting on the stronger predictor ($\beta=0.66$, $p=0.02$), reinforcing the capabilities of larger models.
However, the clear difference is that the intercept ($\beta =2.65$) represents a clear shift from that of the first regression ($\beta=-0.78$), suggesting that as a whole LLMs have improved between mid 2024 and January 2026 at the mathematical task of inferring the more accurate predictor from demonstrated examples.}

\clearpage

\begin{table}[h!]
\caption{Updated regression from Table~\ref{tab:revealed_coefs} using responses from newer models in 2026.}
\vspace{0.5em}
\centering
\label{tab:revealed_coefs_2026}
\begin{tabular}{l|cc}
{{Variable}} &  Coefficient & $p$-Value\\
\toprule
Intercept & $\beta=2.65$ & $p<0.001$ \\
Strong Algorithm & $\beta=1.46$ & $p=0.06$ \\
Complex LLM & $\beta=0.66$ & $p=0.02$ \\
Interaction & $\beta=0.67$ & $p=0.33$ \\
\end{tabular}
\end{table}

\xhdr{Stated-Revealed Comparison}
\update{Figure \ref{fig:stated_revealed_new_models} compares the updated Study 1 and 2 results for the stated-revealed algorithm aversion gap. 
Overall, models are \textit{stating much more preference} and \textit{revealing slightly less preference} towards algorithms as compared to before. 
These shifts have resulted in the \textit{stated-revealed} relative risk of trusting a human $RR_{sr} = P(human|stated)/P(human|revealed)$ to shift towards stating algorithm appreciation relatively more than revealing, which is opposite to the effect we found previously. 
We find that $RR_{sr} < 1$ for all models, most with $p<0.01$ except \texttt{claude-4-haiku} and \texttt{llama-4-scout}. 
Nonetheless, tracking previous results in Section~\ref{sec:comp_results}, the less complex model in each family has a greater $RR_{sr}$ than the more complex model, indicating that model complexity may be correlated with more stated algorithm appreciation.
}

\end{document}